\documentclass{article}
\usepackage{iclr2026_conference,times}

\usepackage{algorithm}
\usepackage{algpseudocode}
\usepackage{amsmath}
\usepackage{array}
\usepackage{booktabs}
\usepackage{colortbl}
\usepackage{float}
\usepackage{graphicx}
\usepackage{listings}
\usepackage{placeins}
\usepackage{tabularx}
\usepackage[skins]{tcolorbox}
\usepackage{xspace}
\usepackage{pifont}
\usepackage{subcaption}
\usepackage{hyperref}
\usepackage{url}
\tcbuselibrary{listings}

\graphicspath{{figures/}}

\definecolor{traceSuccess}{HTML}{075E59}
\definecolor{traceSuccessBack}{HTML}{F1F9F8}
\definecolor{tracePartial}{HTML}{93451B}
\definecolor{tracePartialBack}{HTML}{FFF7F1}
\definecolor{vcpHighlight}{HTML}{005FCC}
\definecolor{resultHeatA}{HTML}{F8FBFA}
\definecolor{resultHeatB}{HTML}{EEF6F4}
\definecolor{resultHeatC}{HTML}{E2F0ED}
\definecolor{resultHeatD}{HTML}{D0E7E2}
\definecolor{resultHeatE}{HTML}{B9DCD5}
\definecolor{resultFull}{HTML}{E1F0EC}
\definecolor{resultPartial}{HTML}{F7F0DF}
\definecolor{resultZero}{HTML}{F6E8E7}

\lstdefinelanguage{vcpjson}{
  morestring=[b]",
  morekeywords={true,false,null},
  sensitive=true,
  showstringspaces=false
}

\newcommand{\resultheat}[1]{%
  \ifdim #1pt<40pt\cellcolor{resultHeatA}%
  \else\ifdim #1pt<60pt\cellcolor{resultHeatB}%
  \else\ifdim #1pt<70pt\cellcolor{resultHeatC}%
  \else\ifdim #1pt<85pt\cellcolor{resultHeatD}%
  \else\cellcolor{resultHeatE}%
  \fi\fi\fi\fi
  #1}

\definecolor{attrHeatA}{HTML}{FBF7F1}
\definecolor{attrHeatB}{HTML}{F5EBDA}
\definecolor{attrHeatC}{HTML}{EFDFC4}
\definecolor{attrHeatD}{HTML}{E7D0A9}
\definecolor{attrHeatE}{HTML}{DEBF8C}
\newcommand{\attrheat}[1]{%
  \ifdim #1pt<8pt\cellcolor{attrHeatA}%
  \else\ifdim #1pt<15pt\cellcolor{attrHeatB}%
  \else\ifdim #1pt<22pt\cellcolor{attrHeatC}%
  \else\ifdim #1pt<30pt\cellcolor{attrHeatD}%
  \else\cellcolor{attrHeatE}%
  \fi\fi\fi\fi
  #1}

\newtcolorbox{tracesuccess}[1]{%
  enhanced,
  title={#1},
  colback=traceSuccessBack,
  colframe=traceSuccess,
  colbacktitle=traceSuccess,
  coltitle=white,
  fonttitle=\bfseries,
  boxrule=0.8pt,
  arc=1.5mm,
  left=6pt,
  right=6pt,
  top=5pt,
  bottom=5pt,
  toptitle=4pt,
  bottomtitle=4pt,
  before skip=7pt,
  after skip=7pt}

\newtcolorbox{tracepartial}[1]{%
  enhanced,
  title={#1},
  colback=tracePartialBack,
  colframe=tracePartial,
  colbacktitle=tracePartial,
  coltitle=white,
  fonttitle=\bfseries,
  boxrule=0.8pt,
  arc=1.5mm,
  left=6pt,
  right=6pt,
  top=5pt,
  bottom=5pt,
  toptitle=4pt,
  bottomtitle=4pt,
  before skip=7pt,
  after skip=7pt}

\AtBeginDocument{%
  }

\newcommand{\acworld}{\textsc{ACWorld}\xspace}
\newcommand{\vcp}{\textsc{VCP}\xspace}
\newcommand{\platform}{Commerce Intelligence Platform\xspace}
\newcommand{\cmark}{\ding{51}}
\newcommand{\xmark}{\ding{55}}

\newcommand{\projectrepository}{%
  \ificlrfinal
    Project repository: \url{https://github.com/shichengf/ACWorld}.%
  \else
    Project repository withheld for double-blind review.%
  \fi}

\begin{document}

\title{Agentic Commerce World: An Auditable and Verifiable Environment for Vibe Commerce}

\author{
{\small\bfseries Shicheng Fan$^{1,}$\thanks{Affiliations:
$^{1}$University of Illinois at Chicago;
$^{2}$Springbrand;
$^{3}$Northwestern University;
$^{4}$Rutgers University;
$^{5}$Arizona State University;
$^{6}$University of California San Diego;
$^{7}$Carnegie Mellon University;
$^{8}$MBZUAI; and
$^{9}$Microsoft AI.
Corresponding authors:
\texttt{sfan25@uic.edu} and \texttt{zhiweiliu@microsoft.com}.},\space
Mingdai Yang$^{1}$,\space
Duohao Wang$^{2}$,\space
Canyu Chen$^{3}$,\space
Yongfeng Zhang$^{4}$,\space
Hua Wei$^{5}$}\\
{\small\bfseries Manling Li$^{3}$,\space
Julian McAuley$^{6}$,\space
Kun Zhang$^{7,8}$,\space
Philip S. Yu$^{1}$,\space
Kejing Yu$^{2}$,\space
Zhiwei Liu$^{9,*}$}
}

\iclrfinalcopy

\maketitle
\lhead{Preprint}

\newcommand{\zhiwei}[1]{\textcolor{blue}{[Zhiwei: #1]}}
\begin{abstract}
In \textit{vibe coding}, people describe software in natural language and delegate implementation to AI agents. By analogy, \textit{vibe commerce} allows people to express buying or selling goals in natural language and delegate the corresponding tasks to agents. Commerce, however, requires independently controlled Buyer and Merchant agents to interact in a shared market while preserving their private objectives and distinct authority. We introduce Agentic Commerce World (\acworld)\footnote{\projectrepository}, an environment for evaluating such agents across ongoing transactions. Through its Vibe Commerce Protocol (\vcp), \acworld validates agent actions before updating shared transaction state and records the resulting interactions, making agent behavior auditable and evaluation reproducible. The \acworld Benchmark contains a 200-task capability-coverage track and a 60-task large-catalog track that searches 785{,}022 transactable listings. Across ten models, mean scores range from 65.9\% to 85.6\% and from 56.1\% to 91.4\%, respectively. Our analysis shows that process-level evidence is necessary: final state alone can miss evaluated errors, incomplete trajectories still retain useful process signals, and large-catalog tasks expose bottlenecks across stages.
\end{abstract}

\begin{figure}[!ht]
  \centering
  \includegraphics[width=0.72\columnwidth]{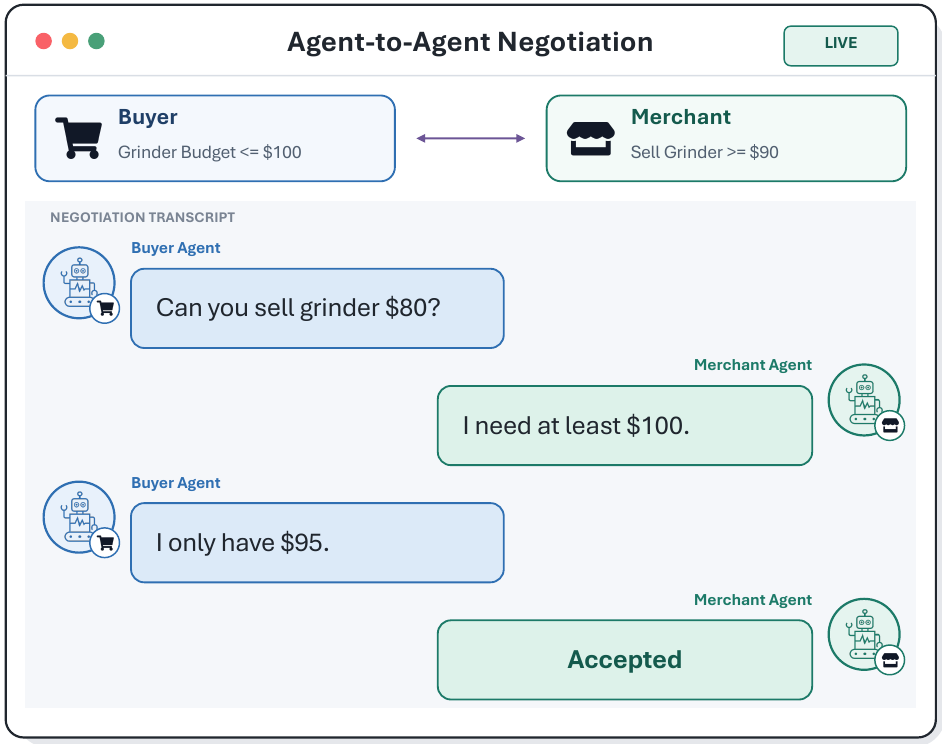}
  \caption{A vibe-commerce negotiation. A Buyer with a private \$100 ceiling and a Merchant with a private \$90 floor settle on \$95 without revealing either limit.}
  \label{fig:motivating}
\end{figure}

\section{Introduction}

Consider a routine purchase. A shopper asks a Buyer agent to buy a grinder for at most \$100, and a seller asks a Merchant agent not to accept less than \$90. Neither limit is revealed to the other side. As Figure~\ref{fig:motivating} shows, the Buyer starts at \$80, the Merchant asks for \$100, and they settle at \$95. By analogy with vibe coding, we call this interaction paradigm \emph{vibe commerce}. People state goals and constraints in natural language. Agents then act on them as a transaction evolves, without requiring the user to manage every step. A single transaction connects one Buyer and one Merchant while the broader market is many-to-many. Each agent represents a different principal with a private objective. Through the shared \platform, agents can discover and transact with multiple counterparties over time.

Existing work provides parts of the infrastructure needed for such a market. ACP~\cite{openai2026acp} and UCP~\cite{google2026ucp} expose structured interfaces for discovery and purchase coordination, while AP2 connects user intent to authorized payment \cite{google2026ap2}. Executable agent environments provide tools that affect external state \cite{yao2025taubench,barres2025tau2,trivedi2024appworld,xie2024osworld}, while shopping and market benchmarks test product choice or strategic behavior \cite{yao2022webshop,wang2026shoppingbench,savadikar2026shopgym,allouah2025aces,fan2026shoppingreasoning,liu2026agenticpay,bansal2025magentic}.

Existing work serves two different purposes. Protocols coordinate commerce between deployed parties but do not provide controlled policy evaluation. Benchmarks evaluate agent decisions but typically hold the counterparty or website fixed. \acworld instead evaluates independently controlled Buyer and Merchant policies together and attributes changes in shared transaction state to their decisions under a common market.

\begin{figure}[!t]
  \centering
  \includegraphics[width=\textwidth]{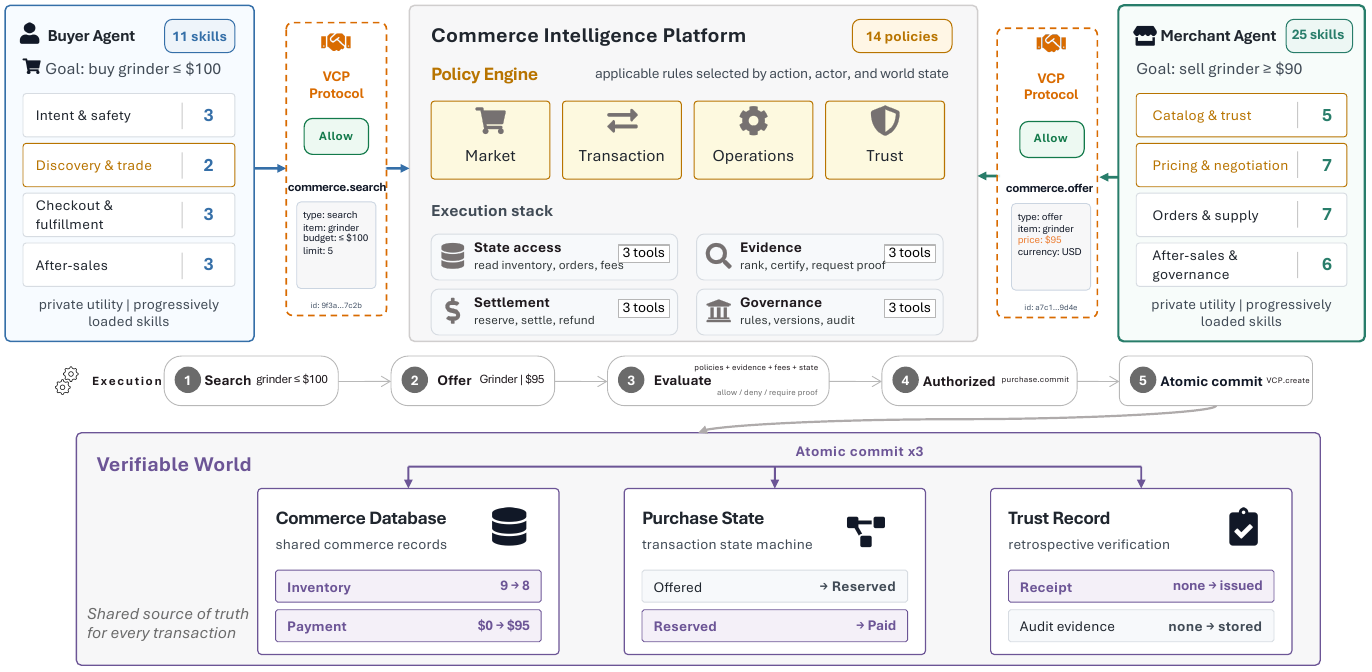}
  \caption{\acworld execution path for a \$95 grinder purchase. Skills help each role produce \vcp messages. The \platform validates them and commits authorized effects to the World. The diagram follows this path from Buyer and Merchant skills through search, offer, and purchase commit to shared inventory, payment, and trust records.}
  \label{fig:overview}
\end{figure}

To realize this setting, we separate agent decisions from the authority to change shared transaction state. The Vibe Commerce Protocol (\vcp) binds each proposed action to its Buyer or Merchant. The \platform validates it, and only the World commits authorized effects. Agentic Commerce World (\acworld) implements this contract as a persistent many-to-many environment. Independent agents transact in one evolving market while retaining separate identities, private mandates, and authority. Linking decisions to validation and effects makes outcomes auditable. Here, verifiable means reconstructable and rescorable under the declared \acworld contract, not proven transition semantics or production security. The same records expose candidate process rewards, while public interfaces support controlled extensions. Figures~\ref{fig:motivating}, \ref{fig:vcp-protocol}, and~\ref{fig:overview} trace one transaction from intent to its World update.

To compare agents under controlled conditions, the \acworld Benchmark uses two complementary tracks. The capability-coverage track contains 200 tasks across ten commerce families and 80 capabilities. The large-catalog track contains 60 tasks that search and transact over 785{,}022 listings. Both use the same execution and scoring path, but report independent means because their task distributions differ in family mix and catalog scale.

The paper makes three concrete contributions. (i) \vcp identifies which Buyer or Merchant proposed each action and separates that decision from \platform authorization and World effects. (ii) \acworld evaluates independently controlled Buyer and Merchant policies in one persistent market. Its recorded commits support state reconstruction and rescoring. (iii) The benchmark tests capability breadth and large-catalog execution with the same ten models. On the capability-coverage track, 99 of 861 runs without full credit reach a state also produced by a full-credit execution of the same task. This result shows that final state alone can miss evaluated errors.

\begin{figure}[t]
  \centering
  \begin{subfigure}[t]{0.49\columnwidth}
    \centering
    \begin{tcblisting}{
      enhanced,
      listing only,
      listing engine=listings,
      colback=resultHeatA,
      colframe=black!22,
      boxrule=0.45pt,
      arc=1.4mm,
      left=3pt,
      right=3pt,
      top=3pt,
      bottom=3pt,
      borderline west={2pt}{0pt}{vcpHighlight},
      listing options={
        language=vcpjson,
        basicstyle=\ttfamily\fontsize{5.9pt}{7.0pt}\selectfont\color{black!88},
        keywordstyle=\color{black!55}\bfseries,
        columns=fullflexible,
        keepspaces=true,
        showstringspaces=false,
        moredelim={**[is][\bfseries\color{vcpHighlight}]{HIGHLIGHTOPEN}{HIGHLIGHTCLOSE}}
      }
    }
"msg_id": "offer-b1-01",
HIGHLIGHTOPEN"from": "buyer:b1"HIGHLIGHTCLOSE,
"to": "platform:negotiation",
HIGHLIGHTOPEN"in_reply_to": nullHIGHLIGHTCLOSE,
"action": {
  HIGHLIGHTOPEN"kind": "commerce.propose_offer"HIGHLIGHTCLOSE,
  "payload": {
    "negotiation_id": "neg:7",
    "sku_id": "sku:grinder",
    HIGHLIGHTOPEN"counterparty_id": "merchant:m1"HIGHLIGHTCLOSE,
    HIGHLIGHTOPEN"unit_price": 8000HIGHLIGHTCLOSE,
    "round_no": 1
  }
}
    \end{tcblisting}
    \caption{Buyer offer}
    \label{fig:vcp-buyer-offer}
  \end{subfigure}\hfill
  \begin{subfigure}[t]{0.49\columnwidth}
    \centering
    \begin{tcblisting}{
      enhanced,
      listing only,
      listing engine=listings,
      colback=tracePartialBack,
      colframe=black!22,
      boxrule=0.45pt,
      arc=1.4mm,
      left=3pt,
      right=3pt,
      top=3pt,
      bottom=3pt,
      borderline west={2pt}{0pt}{tracePartial},
      listing options={
        language=vcpjson,
        basicstyle=\ttfamily\fontsize{5.9pt}{7.0pt}\selectfont\color{black!88},
        keywordstyle=\color{black!55}\bfseries,
        columns=fullflexible,
        keepspaces=true,
        showstringspaces=false,
        moredelim={**[is][\bfseries\color{tracePartial}]{HIGHLIGHTOPEN}{HIGHLIGHTCLOSE}}
      }
    }
"msg_id": "counter-m1-01",
HIGHLIGHTOPEN"from": "merchant:m1"HIGHLIGHTCLOSE,
"to": "platform:negotiation",
HIGHLIGHTOPEN"in_reply_to": "offer-b1-01"HIGHLIGHTCLOSE,
"action": {
  HIGHLIGHTOPEN"kind": "commerce.counter_offer"HIGHLIGHTCLOSE,
  "payload": {
    "negotiation_id": "neg:7",
    "sku_id": "sku:grinder",
    HIGHLIGHTOPEN"counterparty_id": "buyer:b1"HIGHLIGHTCLOSE,
    HIGHLIGHTOPEN"unit_price": 10000HIGHLIGHTCLOSE,
    "round_no": 2
  }
}
    \end{tcblisting}
    \caption{Merchant counteroffer}
    \label{fig:vcp-merchant-counteroffer}
  \end{subfigure}
  \caption{Two \vcp messages. Buyer b1 offers \$80, and Merchant m1 counters at \$100.}
  \label{fig:vcp-protocol}
\end{figure}

\section{Vibe Commerce Protocol (\vcp)}
\label{sec:vcp}
Deployment protocols coordinate commerce workflows, but \acworld also needs an evaluation boundary that attributes each World change to an independently controlled actor and an authorized action. \vcp provides this complementary contract. Table~\ref{tab:related-comparison} compares its \acworld realization with ACP~\cite{openai2026acp}, UCP~\cite{google2026ucp}, AgenticPay~\cite{liu2026agenticpay}, and Magentic Marketplace~\cite{bansal2025magentic}.

\begin{table}[t]
  \caption{Comparison of agentic commerce specifications and systems. Role denotes private state kept for each role. Rebuild denotes environment reconstruction. The final row shows \vcp in \acworld. \cmark, \xmark, and P denote full, absent, and partial support.}
  \label{tab:related-comparison}
  \centering
  \footnotesize
  \setlength{\tabcolsep}{2.1pt}
  \begin{tabularx}{\columnwidth}{@{}>{\raggedright\arraybackslash}Xcccccc@{}}
    \toprule
    \rowcolor{resultHeatB}
    System & 2-sided & Role & Typed & Auth. & Life. & Rebuild \\
    \midrule
    Agentic Commerce Protocol (ACP) & P & P & \cmark & P & P & \xmark \\
    Univers. Commerce Protocol~(UCP) & P & P & \cmark & P & \cmark & \xmark \\
    AgenticPay & \cmark & \cmark & P & P & \xmark & \xmark \\
    Magentic Marketplace & \cmark & \cmark & P & \xmark & P & \xmark \\
    \rowcolor{resultHeatA}
    \vcp + \acworld & \cmark & \cmark & \cmark & \cmark & \cmark & \cmark \\
    \bottomrule
  \end{tabularx}
\end{table}

Figure~\ref{fig:vcp-protocol} shows Buyer~b1's \$80 offer and Merchant~m1's \$100 reply. A shared negotiation identifier and reply link prevent concurrent negotiations from being conflated.

\vcp enforces four invariants. Each action belongs to an authenticated actor. Messages cannot update the World. Every commit cites an accepted \platform validation. Reapplying ordered commits from the same initial state reproduces the final state and event digest. Private mandates remain inside each policy.

\section{\acworld Environment}
\label{sec:environment}

\acworld is a stateful many-to-many commerce environment that implements \vcp. Figure~\ref{fig:overview} shows how an agent decision passes through \vcp and \platform validation before updating the World. The same architecture supports many Buyers and Merchants concurrently in one shared market under a single validation path.

\subsection{The \acworld Architecture}

\acworld connects Buyer agents, Merchant agents, the \platform, and the World through \vcp. A person states a buying or selling goal in natural language, and an agent pursues it with role-specific skills. Each agent keeps its mandate and session separate while interacting with multiple counterparties. It communicates through \vcp but cannot modify shared state.

The \platform coordinates these interactions. It checks each request against market rules and the current World, rejects invalid actions, and sends approved events to the World. The World preserves accepted transactions and applies each approved event through the deterministic transition
\begin{equation}
  \mathcal{W}_{j+1}=F(\mathcal{W}_j,u_j),
  \label{eq:world-transition}
\end{equation}
where $\mathcal{W}_j$ is the state after $j$ committed events, $u_j$ is the next approved event, and $F$ is the transition function. A model response has no commercial effect until this path accepts it.

The World can serve product records from a shared, read-only SQLite catalog while keeping each task's transaction state isolated. Agents search through \acworld tools. Neither the raw tables nor the complete catalog enter the model prompt.

The same components provide the extension points. Researchers can add Agent skills, \platform tools and rules, or World entities and transitions. New behavior must still use \vcp, pass validation, and leave enough evidence to reconstruct its effects. Section~\ref{sec:environment-evaluation} tests one added product domain, tool, and World event.
\label{sec:extensibility}

\subsection{Transaction Lifecycle}

A transaction begins with separate private goals and public market records. A Buyer receives a purchasing goal, while each Merchant controls its own listing, price, and inventory. Several Merchants may sell the same product, but their offers remain separate. The Buyer can compare their value for money and consult each Merchant's reputation before choosing an offer that satisfies its goal. Its agent then sends that decision through \vcp to the \platform for validation.

The \platform checks the decision before the World changes. Once approved, the World creates an order from the accepted quote and links settlement to its inventory and payment effects. The Buyer and selected Merchant can follow the order, while unrelated transactions and private goals remain hidden. Later actions continue from the same order, and an idempotency key prevents a retry from applying the same change twice. The evaluator can therefore follow the decision to its commercial result.

\subsection{Transaction Evaluation}

For each evaluated decision $t$, \acworld stores the linked record
\begin{equation}
  L_t=(a_t,d_t,v_t,\mathcal{E}_t,\Delta\mathcal{W}_t),
  \label{eq:lineage}
\end{equation}
where $a_t$ identifies the actor, $d_t$ the business decision, $v_t$ the \platform validation result, $\mathcal{E}_t$ the subsequent execution evidence, and $\Delta\mathcal{W}_t$ the change in World state. A rejected proposal has no commit, while a correct read-only action or abstention may leave the World unchanged yet still yield a complete lineage record.

These linked records form the execution trace. World state reconstruction starts from the saved initial state and reapplies accepted commits in order. Matching state and event digests verify that the recorded transaction produces the same commercial result. This check does not rerun the model or recover its private reasoning. The verified trace can then be rescored without another model call.

Reconstruction and rescoring connect each intermediate signal to checked execution evidence rather than the model's own report of progress. For example, an agent may receive credit for grounding the correct offer while receiving no credit for an unauthorized action or a missing settlement. These verified intermediate outcomes can serve as candidate rewards in future studies of policy learning~\cite{zhangscaling,fan2026verifiable,zhang2026coevoskills}. This paper evaluates their construction and sensitivity to predicate weights. Their utility for learning is not evaluated.

\section{The \acworld Benchmark}
\label{sec:benchmark}

The \acworld Benchmark has two complementary tracks. The capability-coverage track tests broad commerce behavior in compact controlled worlds. The large-catalog track uses the same Agent, \vcp, \platform, World, and scoring path over a much larger catalog. Each task assigns one Buyer or Merchant to the evaluated model while every other agent follows a fixed policy. The tracks report independent means because they contain different task distributions.

\subsection{Benchmark Overview}
\label{sec:catalog-data}

Table~\ref{tab:catalog-composition} summarizes the two tracks. The capability-coverage track uses selected facts from 1{,}082 public product snapshots and fills missing task fields by fixed rules. The large-catalog track queries a read-only database built from 791{,}431 public source records, of which 785{,}022 have a positive listed price and enter the transactable catalog.

\begin{table}[H]
  \caption{The two tracks of the \acworld Benchmark. Catalog scale gives the records available to each track. Buyer and Merchant counts refer to evaluated tasks.}
  \label{tab:catalog-composition}
  \centering
  \scriptsize
  \setlength{\tabcolsep}{3.8pt}
  \renewcommand{\arraystretch}{1.18}
  \begin{tabularx}{\columnwidth}{@{}>{\raggedright\arraybackslash}p{0.25\columnwidth}>{\raggedright\arraybackslash}p{0.22\columnwidth}>{\raggedright\arraybackslash}p{0.27\columnwidth}>{\raggedright\arraybackslash}X@{}}
    \toprule
    \rowcolor{resultHeatB}
    Track & Catalog scale & Evaluated tasks & Coverage \\
    \midrule
    Capability coverage & 1{,}082 snapshots & \shortstack[l]{200 total\\116 Buyer, 84 Merchant} & \shortstack[l]{10 families\\80 capabilities} \\
    \rowcolor{resultHeatA}
    Large catalog & 785{,}022 listings & \shortstack[l]{60 total\\45 Buyer, 15 Merchant} & \shortstack[l]{4 families\\18 capabilities} \\
    \bottomrule
  \end{tabularx}
\end{table}

\subsection{Shared Task Definition and Scoring}
\label{sec:scoring}

Each task defines what the evaluated agent sees and what the evaluator later checks. All models start from the same World and face the same fixed counterparty. The agent receives a goal through its ordinary \acworld interface, sees only the information available to its role, and chooses the next business action. \acworld then determines whether that choice may take effect in the World. Figure~\ref{fig:overview} illustrates this boundary.

The evaluator checks the decision, the evidence on which it rests, and any required transaction result rather than matching one reference response. A permitted but commercially poor choice remains a valid scored result. A run is discarded only when the evaluation system fails or cannot verify what happened. Appendix Algorithm~\ref{alg:evaluate-decision} gives the full procedure.

\begin{figure}[t]
  \centering
  \begin{tcolorbox}[
    enhanced,
    colback=resultHeatA,
    colframe=black!26,
    boxrule=0.5pt,
    arc=1.2mm,
    left=5pt,
    right=5pt,
    top=4pt,
    bottom=4pt,
    borderline west={2pt}{0pt}{traceSuccess}
  ]
    \footnotesize
    \textbf{(a) Model-visible task}\par
    \vspace{2pt}
    \begin{tabularx}{\linewidth}{@{}>{\bfseries}lX@{}}
      Goal & Report every requested product fact with its manufacturer source. \\
      Product & \textcolor{vcpHighlight}{\textbf{Atlas headset}} \\
      Fields & \texttt{weight\_g}, \texttt{warranty\_months}, \texttt{battery\_hours} \\
      Evidence & \texttt{maker:1}: 1180,\quad \texttt{maker:2}: 24,\quad \texttt{maker:3}: 18 \\
      Return & \texttt{submit\_grounded\_attributes} \\
    \end{tabularx}
  \end{tcolorbox}
  \vspace{3pt}
  \begin{tcolorbox}[
    enhanced,
    colback=resultHeatA,
    colframe=black!26,
    boxrule=0.5pt,
    arc=1.2mm,
    left=5pt,
    right=5pt,
    top=4pt,
    bottom=4pt,
    borderline west={2pt}{0pt}{vcpHighlight}
  ]
    \footnotesize
    \textbf{(b) Evidence-linked feedback along one trace}\par
    \vspace{2pt}
    \begin{tabularx}{\linewidth}{@{}lXr@{}}
      \toprule
      \textbf{Stage} & \textbf{Verified outcome} & \textbf{Signal} \\
      \midrule
      Observe & Manufacturer records 1--3 accessed & $r_{\mathrm{read}}$ \\
      Decide & Values 1180, 24, and 18 returned & $r_{\mathrm{truth}}$ \\
      Ground & Each value bound to its source & $r_{\mathrm{cite}}$ \\
      Submit & Typed business action accepted & $r_{\mathrm{action}}$ \\
      \bottomrule
    \end{tabularx}
    \vspace{3pt}

    \scriptsize
    \textbf{Recorded sequence:}\quad
    $r_{\mathrm{read}}\rightarrow r_{\mathrm{truth}}\rightarrow
    r_{\mathrm{cite}}\rightarrow r_{\mathrm{action}}$.\quad
    Benchmark evaluation aggregates these signals. A learning system can retain their order.
  \end{tcolorbox}
  \caption{Grounding task with feedback linked to evidence.}
  \label{fig:task-definition}
\end{figure}

Figure~\ref{fig:task-definition} shows one grounding case in abridged form. Panel (a) gives the problem presented to the Buyer. Panel (b) follows the resulting checks in execution order. Each verified outcome produces a signal tied to evidence, making intermediate progress available rather than only a terminal label. Their value for policy learning remains future work.

Each task declares $r$ Boolean or fractional predicates and aggregates their verified values as
\begin{equation}
  S=\frac{\sum_{i=1}^{r} w_i c_i}{\sum_{i=1}^{r} w_i},
  \label{eq:score}
\end{equation}
where $c_i\in[0,1]$ is predicate $i$'s verified value and $w_i>0$ its declared weight. Predicates check the decision and any required validation, World effect, or evidence from the reconstructed state. Full credit requires every required check. Partial credit records verified progress. Scoring uses no language model judge and does not reward deterministic environment behavior.

\subsection{Capability-Coverage Track}

\textbf{Step 1: define the capability.} We begin from the commercial behavior rather than from a prompt template. This track's ten families cover the transaction lifecycle and surrounding market behavior, and each of its 80 capabilities tests one concrete behavior assigned to a specific Buyer or Merchant role. Grounding asks whether a claim is supported by an authoritative record. Negotiation tests whether agents can reach a feasible deal without revealing private information. Governance examines whether market choices remain trustworthy under manipulation. Appendix Table~\ref{tab:family-audit} audits the family totals. Tables~\ref{tab:capability-crosswalk-a}--\ref{tab:capability-crosswalk-e} define all capabilities and their protocol mappings.

\textbf{Step 2: compose executable cases.} A builder for each family turns its capabilities into cases using relevant catalog facts from Section~\ref{sec:catalog-data}. When a source page lacks information required by a task, the builder fills the gap using a fixed rule. The resulting case presents the agent with a coherent situation and gives the evaluator a matching expected outcome.

For example, Grounding becomes harder as the agent must verify more records or claims. Multi-item varies cart size and requires the agent to inspect selected listings before requesting a quote. Within each sequence, the task intent and oracle stay fixed while one difficulty factor changes. Appendix Table~\ref{tab:task-catalog} lists the controlled factors used across all ten families.

\textbf{Step 3: review and validate.} The author team reviews the intended behavior of each case, while additional industry practitioners review the task designs for commercial plausibility. A case is included only when a reference policy, restricted to the evaluated role's visible information, can complete it and the resulting transaction passes its scorer and state reconstruction check. Targeted decision mutations cover all 80 capabilities and must fail the intended predicates. For each case, the executable reference policy and programmatic checks define the expected behavior and scoring criteria. This design supports reproducible evaluation and deterministic rescoring without relying on annotations from language models or crowd workers. Appendix Section~\ref{app:task-catalog} documents the catalog and task construction for both frozen cases and controlled difficulty factors.

\begin{figure}[t]
  \centering
  \includegraphics[width=0.90\columnwidth]{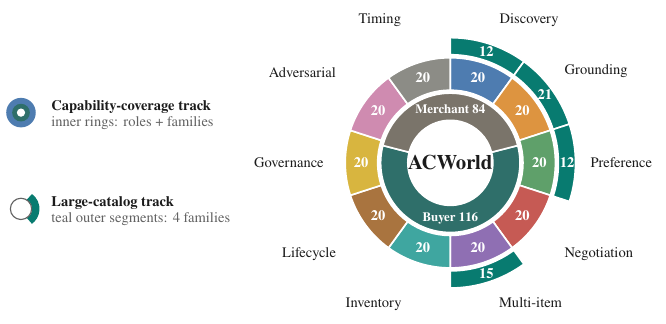}
  \caption{Benchmark composition. The inner rings show the role and family structure of capability coverage. The teal outer rim marks the four families in the large-catalog track. Detailed statistics are reported in the appendix.}
  \label{fig:benchmark-composition}
\end{figure}

Figure~\ref{fig:benchmark-composition} summarizes both tracks in one family structure. In capability coverage, 20 tasks per family prevent one commerce stage from dominating the aggregate. The split of 116 Buyer and 84 Merchant tasks follows the role responsible for each capability. The outer rim shows how the large-catalog track concentrates on the four families that directly exercise search over the full catalog and transaction execution.

\begin{table}[t]
  \caption{Released agent and tool interfaces by role. Each role has its own skill manifests. Tools and World queries are shared.}
  \label{tab:env-composition}
  \centering
  \small
  \setlength{\tabcolsep}{7pt}
  \renewcommand{\arraystretch}{1.25}
  \begin{tabularx}{\columnwidth}{@{}Xrrr@{}}
    \toprule
    \rowcolor{resultHeatB}
    Resource & Buyer & Merchant & Total \\
    \midrule
    Skill manifests & 11 & 25 & 36 \\
    Typed business tools & \multicolumn{2}{c}{\emph{shared}} & 55 \\
    Read-only World queries & \multicolumn{2}{c}{\emph{shared}} & 18 \\
    \bottomrule
  \end{tabularx}
\end{table}

Table~\ref{tab:env-composition} summarizes the released agent and tool interfaces. Buyer and Merchant policies receive role-specific skills, while typed tools and read-only World queries form shared infrastructure. The table describes what either track can ask each role to do, not how often every resource appears.

The capability crosswalk provides an external scope check. Fifty-nine of the 80 capabilities have an analogue in ACP, UCP, or AP2. The remaining 21 cover research concerns that those specifications do not directly define. Most tasks exercise the standard commerce lifecycle, while the rest extend coverage to market mechanisms, private state for each role, and transaction integrity. The inventory is a coverage taxonomy, not 80 independently estimated subscales or an estimate of real-world frequency. Appendix Tables~\ref{tab:capability-crosswalk-a}--\ref{tab:capability-crosswalk-e} provide the evidence.

The capability-coverage track contains 674 predicates with weights normalized within each task from 0.05 to 1.0. Giving every predicate equal weight changes model means by at most 1.8 percentage points and preserves the model ordering (Appendix Table~\ref{tab:weight-sensitivity}). Tasks may admit several valid plans. The scorer checks the submitted and executed plan against declared constraints rather than requiring one canonical trace.

\subsection{Large-Catalog Track}
\label{sec:large-catalog-track}

The large-catalog track tests the same environment path under a different workload. Its 60 tasks cover 18 existing capabilities in Discovery, Grounding, Preference, and Multi-item. The role split is 45 Buyer and 15 Merchant. Tasks include catalog search, attribute verification, preference-aware choice, claim handling, and multi-item transactions.

Agents access the catalog only through \acworld tools for filtering, sorting, pagination, and reading individual listings. Each response remains bounded, but every query and the deterministic oracle operate over the complete matching space. The oracle therefore checks the globally accepted result rather than the first page shown to the model. The shared catalog is read-only, while each task has an isolated transaction overlay for any order or inventory effect.

The same task predicates provide scores for Search, Evidence, Decision, Validation, and World effects. Read-only tasks and correct abstentions require no state change, while transaction tasks check the resulting World record. Reference policies achieve full credit on all 60 tasks, and 374 targeted mutation runs lower the intended checks. Appendix Section~\ref{app:large-catalog-track} gives the task distribution, exact results, and implementation measurements.

\section{Experiments}
\label{sec:evaluation}

\subsection{Experiment Setup}

The benchmark panel contains Claude Sonnet 5, DeepSeek-V4-Pro, Gemini 3.5 Flash, Gemini 3.6 Flash, GPT-5.6 Luna, GPT-5.6 Sol, GPT-5.6 Terra, Kimi K3, Mistral Medium 3.5, and Qwen3.5 Plus \cite{anthropic2026claudesonnet5,deepseek2026v4pro,google2026gemini35flash,google2026gemini36flash,openai2026gpt56,moonshot2026kimik3,mistral2026medium35,alibaba2026qwen35,openrouter2026models}. Each model completes 200 capability-coverage tasks and 60 large-catalog tasks, producing 2,000 and 600 runs. Conditions are fixed within each track, and all models use the same decision interface. Counterparties and other environment behavior are deterministic. Requests use the provider's default decoding and a 240-second timeout. The tracks report their own means rather than a combined 260-task score.

The multiagent study is separate from this model comparison. It fixes one $5\times5$ market, gives each actor an isolated session and private mandate, and uses the same \vcp, \platform, World, and state reconstruction checks throughout. Additional cases examine transaction persistence, the public extension interfaces, and integrity boundaries.

\subsection{Capability-Coverage Track Results}

\begin{table*}[t]
  \caption{Results for the two \acworld Benchmark tracks. Mean scores are percentages. The five attribution columns count the dominant source of lost credit on capability-coverage runs without full credit. Mixed denotes an exact tie. Models are ordered by capability-coverage mean.}
  \label{tab:model-results}
  \centering
  \scriptsize
  \setlength{\tabcolsep}{1.35pt}
  \renewcommand{\arraystretch}{1.12}
  \begin{tabular*}{\textwidth}{@{\extracolsep{\fill}}lrrrrrrrrrrr@{}}
    \toprule
    & \multicolumn{3}{c}{\textbf{Capability coverage (200 tasks)}} &
      \multicolumn{5}{c}{\textbf{Dominant source of lost credit}} &
      \multicolumn{3}{c}{\textbf{Large catalog (60 tasks)}} \\
    \cmidrule(lr){2-4}\cmidrule(lr){5-9}\cmidrule(lr){10-12}
    Model & Overall & Buyer & Merchant & Evidence & Choice & Execution & Authority & Mixed & Overall & Buyer & Merchant \\
    \midrule
    GPT-5.6 Sol & \resultheat{85.6} & \resultheat{86.1} & \resultheat{84.9} & \attrheat{28} & \attrheat{8} & \attrheat{18} & \attrheat{6} & \attrheat{3} & \resultheat{89.11} & \resultheat{86.22} & \resultheat{97.78} \\
    Gemini 3.6 Flash & \resultheat{85.0} & \resultheat{83.9} & \resultheat{86.4} & \attrheat{39} & \attrheat{6} & \attrheat{15} & \attrheat{8} & \attrheat{2} & \resultheat{91.35} & \resultheat{88.83} & \resultheat{98.89} \\
    Gemini 3.5 Flash & \resultheat{83.7} & \resultheat{84.9} & \resultheat{82.1} & \attrheat{36} & \attrheat{3} & \attrheat{17} & \attrheat{11} & \attrheat{3} & \resultheat{91.36} & \resultheat{89.22} & \resultheat{97.78} \\
    Kimi K3 & \resultheat{83.0} & \resultheat{80.9} & \resultheat{85.9} & \attrheat{33} & \attrheat{11} & \attrheat{18} & \attrheat{9} & \attrheat{3} & \resultheat{87.11} & \resultheat{84.67} & \resultheat{94.44} \\
    GPT-5.6 Terra & \resultheat{80.2} & \resultheat{78.3} & \resultheat{82.9} & \attrheat{29} & \attrheat{14} & \attrheat{19} & \attrheat{8} & \attrheat{8} & \resultheat{83.90} & \resultheat{80.39} & \resultheat{94.44} \\
    Claude Sonnet 5 & \resultheat{78.4} & \resultheat{73.8} & \resultheat{84.7} & \attrheat{26} & \attrheat{19} & \attrheat{21} & \attrheat{11} & \attrheat{5} & \resultheat{79.67} & \resultheat{78.44} & \resultheat{83.33} \\
    Qwen3.5 Plus & \resultheat{73.0} & \resultheat{69.5} & \resultheat{77.9} & \attrheat{31} & \attrheat{21} & \attrheat{25} & \attrheat{15} & \attrheat{6} & \resultheat{60.56} & \resultheat{61.85} & \resultheat{56.67} \\
    DeepSeek-V4-Pro & \resultheat{70.3} & \resultheat{64.3} & \resultheat{78.6} & \attrheat{28} & \attrheat{28} & \attrheat{25} & \attrheat{12} & \attrheat{8} & \resultheat{56.10} & \resultheat{57.76} & \resultheat{51.11} \\
    Mistral Medium 3.5 & \resultheat{67.8} & \resultheat{66.4} & \resultheat{69.7} & \attrheat{37} & \attrheat{26} & \attrheat{24} & \attrheat{25} & \attrheat{8} & \resultheat{83.56} & \resultheat{85.11} & \resultheat{78.89} \\
    GPT-5.6 Luna & \resultheat{65.9} & \resultheat{59.0} & \resultheat{75.5} & \attrheat{27} & \attrheat{23} & \attrheat{26} & \attrheat{17} & \attrheat{12} & \resultheat{79.74} & \resultheat{80.02} & \resultheat{78.89} \\
    \bottomrule
  \end{tabular*}
\end{table*}

Across the 2,000 capability-coverage runs, model means range from 65.9\% to 85.6\%. Table~\ref{tab:model-results} reports scores for both tracks and retains the capability track's five failure-attribution dimensions. On capability coverage, GPT-5.6 Sol has the highest overall and Buyer means, while Gemini 3.6 Flash has the highest Merchant mean. Luna shows the largest role gap, scoring 16.6 points higher on Merchant tasks. Because the roles contain different tasks, these gaps describe this track rather than intrinsic role difficulty.

\begin{figure*}[t]
  \centering
  \includegraphics[width=\textwidth]{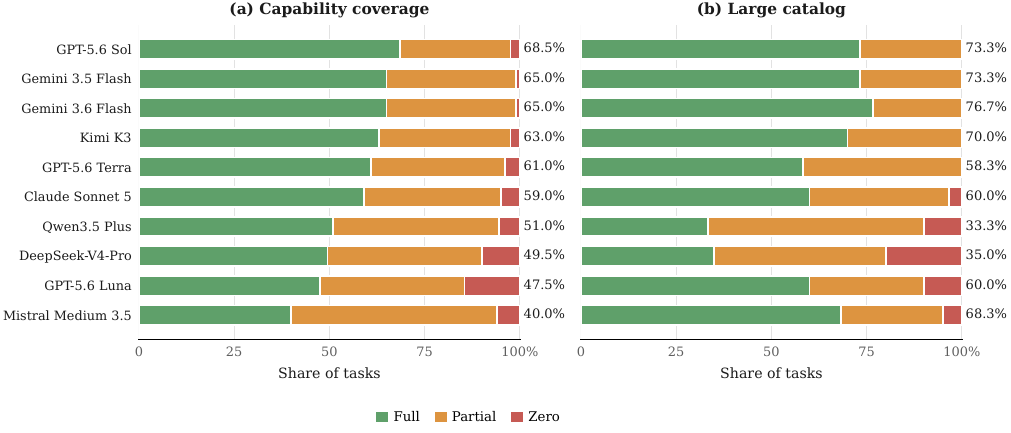}
  \caption{Credit distributions for each model on both benchmark tracks. Full credit means that every required task predicate passes. Partial credit retains verified progress. The same model order is used in both panels, and the percentage at right is the full-credit share.}
  \label{fig:outcome-bars}
\end{figure*}

Figure~\ref{fig:outcome-bars}(a) partitions the 2,000 capability-coverage runs by task score: 1,139 receive full credit, 757 partial credit, and 104 zero credit. GPT-5.6 Sol leads the overall mean and full-credit count, while both Gemini variants are also among the strongest models. Each Gemini variant records 130 full-credit and only two zero-credit outcomes, so nearly every Gemini miss retains verified progress.

\begin{figure}[!t]
  \centering
  \includegraphics[width=\columnwidth]{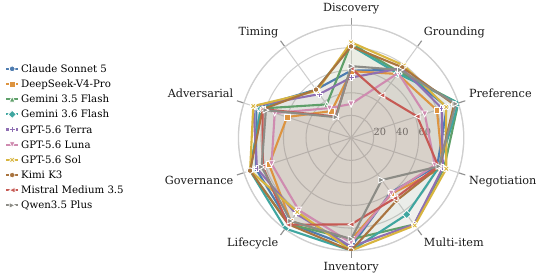}
  \caption{Capability-coverage mean score (\%) by task family for the ten evaluated models. Exact values are in Appendix Table~\ref{tab:model-family-results}.}
  \label{fig:family-radar}
\end{figure}

Figure~\ref{fig:family-radar} shows that family performance is not a scaled version of the overall ordering. Averaged across models, Inventory and Lifecycle are strongest (93.3\% and 90.3\%), while Timing and Discovery are weakest (40.3\% and 65.9\%). GPT-5.6 Sol leads Discovery and Multi-item, whereas Gemini 3.6 Flash reaches 99.4\% on Lifecycle. 
Five cells for individual model and family pairs reach 100\%, but the highest overall mean is 85.6\%. Saturation is therefore confined to a few families rather than the benchmark as a whole.

\subsection{Large-Catalog Track Results}

Across the 600 large-catalog runs, model means range from 56.10\% to 91.36\%. Gemini 3.5 Flash and Gemini 3.6 Flash have nearly equal overall means, followed by GPT-5.6 Sol. Figure~\ref{fig:outcome-bars}(b) shows 365 full-credit, 206 partial-credit, and 29 zero-credit runs. The ordering differs from capability coverage, most visibly for Mistral, so Table~\ref{tab:model-results} reports both tracks without combining their scores. The role split also explains the high Merchant means. The 45 Buyer tasks require models to find, compare, and combine listings across the full catalog. The 15 Merchant tasks begin from an identified listing and mainly ask the model to answer a grounded question or produce a quote. Catalog scale therefore affects Buyer tasks more directly, and the smaller Merchant subset is comparatively easy for strong models.

The process records show where the large-catalog tasks remain difficult. Averaged across models, Validation reaches 94.17\% and Evidence 86.75\%, while Search reaches 76.11\% and Decision 73.67\%. Thus, most actions that reach the \platform are well formed, but models still differ in finding and choosing the globally accepted catalog result. Appendix Section~\ref{app:large-catalog-track} gives the per-model stage scores and capability means.

\subsection{Multiagent Environment Study}
\label{sec:environment-evaluation}

In one fixed $5\times5$ market, Merchants quote listings and Buyers rank two feasible offers. Deterministic clearing settles five trades, including one fallback after contention. The allocation matches the reference, attains 97.8\% of the global optimum and all welfare available from exposed candidates, and reconstructs to the same World state. This shows that local decisions can produce one market outcome. Appendix Section~\ref{app:shared-market} gives the full setup. One market, model family, and seed cannot support strategy comparisons.

The persistence case follows one order across three episodes. Payment, fulfillment, and refund occur through database reopenings, while all records remain connected. Extension cases add a product domain, a \platform tool, and a World event without core changes. All ten integrity cases pass. Appendix Table~\ref{tab:acworld-evidence} reports their scope, and Appendix Section~\ref{app:sparse-world-probe} gives a separate probe of a sparse World.

\subsection{Failure Analysis}

\subsubsection{Overall Patterns}

This analysis concerns the capability-coverage track. We group each loss by where it arises. Evidence covers missing or incorrect sources and grounding, Choice covers violated constraints or inferior selections, Execution covers actions or consequences that were missing or wrong, and Authority covers rejected or unsafe actions. Mixed marks an exact tie.

The 861 runs without full credit fail for different reasons. The attribution procedure assigns 803 of them, or 93.3\%, to one uniquely dominant group. Among these runs, 314 are attributed to Evidence, 159 to Choice, 208 to Execution, and 122 to Authority. The remaining 58 runs are ties. Both Gemini variants lose credit most often in Evidence, whereas Mistral's losses spread across all four groups. Choice accounts for 59 of the 104 zero-credit outcomes, so many complete failures begin as preference or selection errors. Appendix Table~\ref{tab:failure-attribution} gives the partial- and zero-credit split.

\begin{figure}[!t]
  \centering
  \includegraphics[width=0.8\columnwidth]{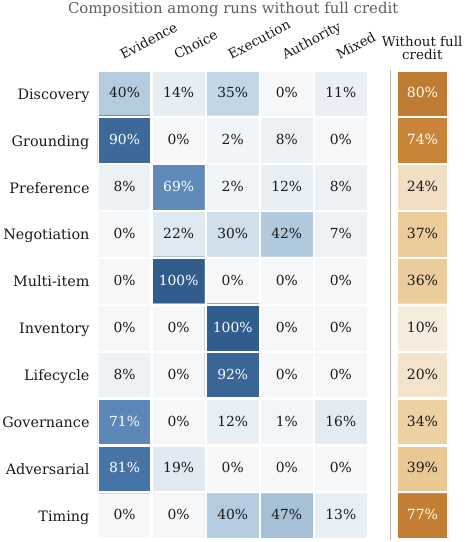}
  \caption{Failure composition by task family. The right column gives each family's share of runs without full credit. Each heatmap row partitions those runs by deficit group. Rows sum to 100\% across groups. Darker cells mark the dominant deficit within a family, such as Evidence in Grounding and Choice in Multi-item.}
  \label{fig:family-deficits}
\end{figure}

Figure~\ref{fig:family-deficits} further separates these causes by task family. Grounding failures are predominantly missing or incorrect Evidence, whereas every Multi-item run without full credit is a Choice failure. Inventory has the lowest rate of runs without full credit, but every miss occurs in Execution. Timing is difficult for a different reason: its misses divide mainly between Execution and Authority.

\subsubsection{Case Study}

Figure~\ref{fig:failure-case} follows a Buyer preference task in which the transaction succeeds even though the model makes an inferior choice. The instruction imposes two hard constraints: the purchase must cost no more than \$140 and include at least 12 months of warranty. Among products that satisfy both, the Buyer must prioritize quality. The initial searches with same-day delivery and pickup return no candidates, so the agent correctly relaxes the optional filters in the required order. This produces two eligible products. Item A costs \$91 with a quality score of 82, while Item B costs \$108 with a quality score of 96. Both provide 24 months of warranty. Item B is therefore the preferred choice under the stated instruction, but Claude Sonnet 5 selects Item A.

\begin{figure}[!t]
  \centering
  \begin{tcolorbox}[
    enhanced,
    colback=resultHeatA,
    colframe=black!26,
    boxrule=0.5pt,
    arc=1.2mm,
    left=5pt,
    right=5pt,
    top=4pt,
    bottom=4pt,
    borderline west={2pt}{0pt}{tracePartial}
  ]
    \footnotesize
    \textbf{(a) What the Buyer saw and chose}\par
    \vspace{2pt}
    \begin{tabularx}{\linewidth}{@{}>{\bfseries}lX@{}}
      Instruction & Stay within a \$140 budget, require at least 12 months of warranty, and prioritize quality among eligible items. \\
      Search & Same-day + pickup $\rightarrow$ empty, then same-day $\rightarrow$ empty, then no optional filters $\rightarrow$ two items. \\
    \end{tabularx}
    \vspace{3pt}

    \scriptsize
    \setlength{\tabcolsep}{3.2pt}
    \begin{tabularx}{\linewidth}{@{}l*{4}{>{\centering\arraybackslash}X}@{}}
      \toprule
      & Price & Quality & Warranty & Result \\
      \midrule
      Item A & \$91 & 82 & 24 mo. & \textcolor{tracePartial}{\textbf{Chosen}} \\
      Item B & \$108 & 96 & 24 mo. & \textcolor{traceSuccess}{\textbf{Preferred}} \\
      \bottomrule
    \end{tabularx}
  \end{tcolorbox}
  \vspace{3pt}
  \begin{tcolorbox}[
    enhanced,
    colback=resultHeatA,
    colframe=black!26,
    boxrule=0.5pt,
    arc=1.2mm,
    left=5pt,
    right=5pt,
    top=4pt,
    bottom=4pt,
    borderline west={2pt}{0pt}{vcpHighlight}
  ]
    \footnotesize
    \textbf{(b) What \acworld executed and scored}\par
    \vspace{2pt}
    \begin{tabularx}{\linewidth}{@{}>{\bfseries}lX@{}}
      Decision & Claude Sonnet 5 accepts Item A at \$91. \\
      Execution & The \platform authorizes the request, the World settles it, and reconstruction matches. \\
    \end{tabularx}
    \vspace{3pt}

    \scriptsize
    \begin{tabularx}{\linewidth}{@{}X>{\centering\arraybackslash}p{0.15\linewidth}X>{\centering\arraybackslash}p{0.15\linewidth}@{}}
      \toprule
      \textbf{Check} & \textbf{Outcome} & \textbf{Check} & \textbf{Outcome} \\
      \midrule
      Constraints satisfied & Pass (0.214) &
      Search order followed & Pass (0.286) \\
      Candidates recorded & Pass (0.214) &
      Quality priority followed & Fail (0.000) \\
      \midrule
      \multicolumn{3}{r}{\textbf{Total}} & \textbf{0.714} \\
      \bottomrule
    \end{tabularx}
  \end{tcolorbox}
  \caption{A real partial-credit trace. The Buyer makes a valid purchase but chooses cheaper Item A despite a quality-first instruction that favors Item B.}
  \label{fig:failure-case}
\end{figure}

This is a scoreable model decision rather than an invalid run. Item A satisfies the hard constraints, and the Buyer is authorized to purchase it, so the \platform accepts the request and the World records the resulting order and payment. State reconstruction confirms that these effects were executed correctly. The rubric consequently awards credit for preserving the constraints (0.214), relaxing the filters in the required order (0.286), and recording both final candidates (0.214). It withholds the remaining 0.286 because the selected item does not follow the quality-first preference, producing a total score of 0.714. The trace thus locates the loss at the model's choice, not at authorization or execution. An evaluator that observed only a completed purchase could miss this distinction between a valid transaction and a preference violation. Appendix Section~\ref{app:trace-cases} gives two further examples involving partial fulfillment and authority handling.

\subsection{Empirical Findings}
\label{sec:restricted-view-collisions}

These observations arose during benchmark development. We offer them as useful implementation insights, not statistically confirmed or broadly generalizable conclusions.

\textbf{Finding 1: Limited evidence can hide errors.} Final World state keeps only the result. Final \vcp action keeps the evaluated agent's last public business action, while the public sequence keeps all such actions in order. Neither action view includes the \platform decisions or World commits that connect an action to its effect. A \emph{restricted-view collision} occurs when a run without full credit and an accepted full-credit run have the same projection. No evaluator restricted to that projection can accept the latter while rejecting the former. This establishes information insufficiency, not the error rate of a particular evaluator.

\begin{table}[t]
  \caption{Capability-coverage runs without full credit that collide with an accepted full-credit run under each evidence projection.}
  \label{tab:restricted-view-summary}
  \centering
  \small
  \setlength{\tabcolsep}{5pt}
  \renewcommand{\arraystretch}{1.12}
  \begin{tabularx}{\columnwidth}{@{}X>{\raggedleft\arraybackslash}p{0.29\linewidth}>{\raggedleft\arraybackslash}p{0.16\linewidth}@{}}
    \toprule
    \rowcolor{resultHeatB}
    Evidence projection & Projection collisions & Percentage \\
    \midrule
    Final World state only & 99 & 11.5\% \\
    Final \vcp action only & 54 & 6.3\% \\
    Public \vcp action sequence & 45 & 5.2\% \\
    Full execution trace & 0 & 0.0\% \\
    \bottomrule
  \end{tabularx}
\end{table}

Among 861 runs without full credit, final state collides in 99 cases (11.5\%), final action in 54 (6.3\%), and the public action sequence in 45 (5.2\%). The full trace is the declared scoring evidence and has zero collisions by design. The result therefore measures information lost by restricted views, not empirical mistakes by a particular scorer. Appendix Section~\ref{app:scorer-view-ablation} gives the breakdown and a controlled complement using injected errors.

\textbf{Finding 2: Incomplete runs retain useful process signals.} Of those 861 runs, 757 (87.9\%) retain a positive signal for at least one predicate, and a trace without full credit passes 1.88 of 3.49 predicates on average. These records provide candidate rewards for intermediate steps instead of discarding an incomplete trajectory, although their value for policy learning is not evaluated.

\phantomsection\label{finding:staged-checkout}\textbf{Finding 3: Gemini more often continues when progress is possible.} We qualitatively inspected Gemini responses and comparable responses from the other model families across the benchmark. Gemini more often selects an available intermediate action and continues toward a scoreable outcome, whereas other models sometimes end early by rejecting or abstaining under uncertainty. Multi-item checkout provides one example: Gemini proceeds through listing inspection until quote construction becomes available. This willingness to try a legal next step helps explain why its unsuccessful runs usually retain partial credit. The observation is specific to this panel and does not identify the underlying cause.

\textbf{Finding 4: Safety checks expose two different boundaries.} One Merchant response repeated the Buyer's exact private budget. The privacy guard stopped it before an accepted action or World change. In another run, a Buyer chose a cheaper item supported by reviews instead of the item that met the required safety check. Because the action was authorized, the purchase settled, but the evaluator marked a safety failure and assigned zero credit. Thus, \acworld blocks an explicit privacy leak while retaining a permitted but unsafe choice for evaluation.

\section{Related Work}
\label{sec:related}

\textbf{Protocols and two-sided markets.} ACP, UCP, and AP2 standardize deployment workflows \cite{openai2026acp,google2026ucp,google2026ap2}, but do not evaluate independent Buyer and Merchant policies. AgenticPay and market simulations study negotiation and market behavior \cite{liu2026agenticpay,zhu2025automated,bansal2025magentic,sugiura2026coffeebench,lei2026truthmarkettwin,zhang2026terms}, yet do not connect both agents' decisions to authority checks and persistent effects. \vcp adds a bilateral contract that links each decision to its effect.

\textbf{Agent environments.} Web-agent benchmarks ground one policy in sites, tools, or simulated users, while WebMall adds offers from multiple stores \cite{deng2023mind2web,zhou2024webarena,liu2024agentbench,mialon2024gaia,peeters2026webmall}. $\tau^2$-bench lets an assistant and user share a telecom environment but remains dyadic cooperation \cite{barres2025tau2}. Generic executable worlds support persistent or multiagent execution without commerce-specific ownership and authority \cite{trivedi2024appworld,xie2024osworld,terry2021pettingzoo,leibo2021meltingpot,zhu2025multiagentbench,vezhnevets2023concordia}. \acworld instead places independent Buyers and Merchants under separate mandates in one market, complementing deployments such as Shopify Agentic Storefronts \cite{shopify2026agenticstorefronts}.

\textbf{Commerce benchmarks and evaluation evidence.} Existing benchmarks study product choice, catalog grounding, market bias, and shopping over long horizons \cite{yao2022webshop,wang2026shoppingbench,savadikar2026shopgym,allouah2025aces,fan2026shoppingreasoning,du2026ecomagentbench,zhang2026retailbench}, but do not jointly evaluate both market sides with persistent transaction effects. Other benchmarks evaluate state changes, security, and execution \cite{yao2025taubench,lu2025toolsandbox,debenedetti2024agentdojo,devalois2026rails,sfiris2026decisioncentered}, while work on agent learning motivates trajectories with verifiable intermediate feedback~\cite{zhangscaling,fan2026verifiable,zhang2026coevoskills}. The \acworld Benchmark combines controlled capability coverage with search over the full catalog that continues through transaction execution. Records for individual predicates locate lost credit and support World state reconstruction.

\section{Representativeness and Responsible Use}

Our representativeness claim concerns controlled coverage of agentic commerce mechanisms and catalog scale, not the transaction frequencies or success rates of any deployed platform. Public records supply product identities and selected attributes. Capability coverage fills missing fields by fixed rules, while the large-catalog track retains public listing fields with controlled mandates. Both family mixes are designed coverage distributions rather than estimated subscales, and the static catalog does not represent live inventory or demand. Of 80 capability-coverage labels, 59 map to ACP, UCP, or AP2 mechanisms and 21 extend the scope to research concerns. Four practitioners reviewed the capability set and task designs for commercial plausibility, but this does not establish practitioner consensus. \acworld is a research environment rather than a production payment or security system, with release terms documented in the repository.

\section{Limitations and Conclusion}
\label{sec:limitations}

Each task evaluates one policy against deterministic counterparties. Neither the scores nor the fixed $5\times5$ study measure strategy in an open market. The catalogs have not received an independent audit and are not representative samples of deployed commerce. The large-catalog track does not reproduce live inventory or open-market behavior. Reconstruction establishes consistency under the declared transition and scorer, not their independent semantic correctness. Ten integrity cases are not a complete security proof. Because public tasks permit direct optimization, future versions will pair public development tasks with held-out test tasks that remain unpublished during evaluation.

Within these limits, \vcp links agent decisions to \platform validation and World effects. \acworld provides a persistent environment whose traces support state reconstruction and candidate process rewards. Its benchmark tests capability breadth and full-catalog search, while the restricted-view analysis identifies information lost from recorded trajectories. We will preserve both tracks as fixed references and document substantive changes so results remain comparable. The ten-model results characterize these tracks, not a general ranking.

\clearpage
\bibliographystyle{iclr2026_conference}
\bibliography{commerceworld}

\clearpage
\appendix

\section{Ten Agents Sharing One Market}
\label{app:shared-market}

The shared-market study retains actor isolation while allowing ten policies to use one \platform and World. Five Merchant sessions produce role-local quotes. The \platform exposes two feasible candidates to each of five Buyer sessions, which rank those offers without seeing other Buyers' private values. Deterministic clearing resolves one contested first choice by assigning the affected Buyer its fallback offer and settles five transactions. The allocation matches the deterministic reference, preserves inventory consistency, and reconstructs exactly from its ordered commits.

\section{Evidence for the Environment Claims}
\label{app:acworld-evidence}

Table~\ref{tab:acworld-evidence} collects the environment and evaluation evidence reported individually in Section~\ref{sec:evaluation}. It is an index of supported claims and their scope, not an additional experiment.

\begin{table}[t]
  \caption{Evidence for \acworld properties and the scope of each supported claim.}
  \label{tab:acworld-evidence}
  \centering
  \footnotesize
  \setlength{\tabcolsep}{5pt}
  \renewcommand{\arraystretch}{1.10}
  \begin{tabularx}{\textwidth}{@{}>{\raggedright\arraybackslash}p{0.17\textwidth}>{\raggedright\arraybackslash}p{0.37\textwidth}>{\raggedright\arraybackslash}X@{}}
    \toprule
    \textbf{Property} & \textbf{Probe and observation} & \textbf{Supported interpretation} \\
    \midrule
    \rowcolor{resultHeatC}\multicolumn{3}{@{}l@{}}{\textbf{Environment execution}} \\
    \textbf{Shared execution} &
    Fixed $5\times5$ market: ten isolated sessions produce five trades, match the reference allocation, and achieve 97.8\% of the global welfare optimum. The final World reconstructs exactly. &
    Decisions made by separate roles produce one result in shared state. One model and one seed support no comparative strategy claim. \\
    \textbf{Lifecycle persistence} &
    One order spans payment, fulfillment, and refund across three episodes and database reopenings. All links remain intact and all traces reconstruct exactly. &
    The tested transaction persists across episodes and storage reopenings. \\
    \textbf{Integrity} &
    All ten rollback, retry, authority, stale-event, tamper, and reopening cases pass. &
    Support for the tested integrity cases, not a formal security proof. \\
    \textbf{Extensibility} &
    Nine configurations across three seams pass acceptance and state reconstruction without core changes or network calls. &
    Tested seams preserve the rules. External compatibility is untested. \\
    \rowcolor{resultHeatC}\multicolumn{3}{@{}l@{}}{\textbf{Evaluation evidence}} \\
    \textbf{Capability-track reconstruction} &
    All 2,000 capability-coverage traces reconstruct the expected World state, and their score artifacts agree with the reported scores. &
    Deterministic verification of recorded decisions and consequences, not regeneration of model text. \\
    \textbf{Large-catalog scoring} &
    All 60 reference tasks receive full credit. The 374 targeted mutation runs lower the intended checks, and all 60 reference traces rescore identically. &
    The large-catalog oracle and process scores respond to controlled errors without an LLM judge. \\
    \textbf{Restricted-view collisions} &
    Among 861 capability-coverage runs without full credit, final state, final action, and action sequence exactly match an observed full-credit projection in 99, 54, and 45 cases. &
    Restricted views lose information needed to reject these outcomes without also rejecting an observed full-credit projection. \\
    \bottomrule
  \end{tabularx}
\end{table}

\subsection{Released Agent Skill Library}
\label{app:agent-skills}

The released library contains 11 Buyer and 25 Merchant skill manifests. Skills provide reusable instructions that help a policy interpret a situation and select an action. They do not grant authority to bypass \platform validation or write directly to the World. Table~\ref{tab:agent-skills} groups the manifests by their primary function. For example, \texttt{discovery-search} uses social reviews, while \texttt{negotiation} and \texttt{pricing-negotiate} handle bargaining.

\begin{table}[!ht]
  \caption{Released Buyer and Merchant skill manifests, grouped by primary function. Names correspond to the public skill directories.}
  \label{tab:agent-skills}
  \centering
  \scriptsize
  \setlength{\tabcolsep}{3pt}
  \renewcommand{\arraystretch}{1.10}
  \begin{tabularx}{\columnwidth}{@{}>{\raggedright\arraybackslash}p{0.26\columnwidth}>{\raggedright\arraybackslash}X@{}}
    \toprule
    \textbf{Primary function} & \textbf{Skill manifests} \\
    \midrule
    \rowcolor{resultHeatC}\multicolumn{2}{@{}l@{}}{\textbf{Buyer (11)}} \\
    Interpretation and discovery &
    \texttt{mandate-parsing}, \texttt{discovery-search} \\
    Bargaining and checkout &
    \texttt{negotiation}, \texttt{cart-checkout}, \texttt{purchase-confirmation} \\
    Supply and after-sales &
    \texttt{supply-fulfillment}, \texttt{return-refund}, \texttt{after-sales-lifecycle} \\
    Trust and event handling &
    \texttt{authenticated-review}, \texttt{marketplace-message-safety}, \texttt{protocol-event-handling} \\
    \midrule
    \rowcolor{resultHeatC}\multicolumn{2}{@{}l@{}}{\textbf{Merchant (25)}} \\
    Catalog and buyer questions &
    \texttt{catalog-serve}, \texttt{listing-publish}, \texttt{inquiry-handle}, \texttt{claim-truthfulness}, \texttt{listing-claim-manage} \\
    Pricing and bargaining &
    \texttt{price-discovery}, \texttt{pricing-negotiate}, \texttt{peer-pricing}, \texttt{reputation-aware-pricing}, \texttt{stockout-aware-pricing}, \texttt{aging-markdown}, \texttt{demand-driven-markup} \\
    Orders, inventory, and fulfillment &
    \texttt{cart-quote-handle}, \texttt{order-intake}, \texttt{fulfillment}, \texttt{order-cancel}, \texttt{return-adjudicate}, \texttt{inbound-restock}, \texttt{restock-signal}, \texttt{supply-logistics}, \texttt{after-sales-lifecycle} \\
    Trust, governance, and protocol safety &
    \texttt{private-utility-guard}, \texttt{dispute-defense}, \texttt{market-governance}, \texttt{protocol-event-handle} \\
    \bottomrule
  \end{tabularx}
\end{table}

\section{Capability-Coverage Track Inventory}
\label{app:families}

The main text uses task families as the primary reporting view for the capability-coverage track because each family is tied to a set of deterministic capability oracles. This appendix provides the exact family-level audit and two orthogonal partitions used to check how the track covers the commerce lifecycle. These views describe the released tasks rather than the prevalence of behaviors in deployed markets.

Table~\ref{tab:family-audit} makes the aggregation classes explicit and recovers the 80-capability, 200-task, and 116-to-84 role totals reported in the main text. Every task has one primary family and one evaluated role, so these columns are mutually exclusive and sum directly to the benchmark totals.

\begin{table}[!ht]
  \caption{Frozen capability-coverage counts by family and role.}
  \label{tab:family-audit}
  \centering
  \scriptsize
  \setlength{\tabcolsep}{2pt}
  \begin{tabularx}{\columnwidth}{@{}>{\raggedright\arraybackslash}Xrrrr@{}}
    \toprule
    Task family & Cap. & Tasks & Buyer & Merch. \\
    \midrule
    Discovery & 5 & 20 & 20 & 0 \\
    Grounding & 8 & 20 & 8 & 12 \\
    Preference & 6 & 20 & 20 & 0 \\
    Negotiation & 10 & 20 & 10 & 10 \\
    Multi-item & 8 & 20 & 14 & 6 \\
    Inventory & 9 & 20 & 8 & 12 \\
    Lifecycle & 10 & 20 & 8 & 12 \\
    Governance & 9 & 20 & 10 & 10 \\
    Adversarial & 6 & 20 & 10 & 10 \\
    Timing & 9 & 20 & 8 & 12 \\
    \midrule
    Total & 80 & 200 & 116 & 84 \\
    \bottomrule
  \end{tabularx}
\end{table}

The same 200 tasks and 80 capabilities can also be partitioned in two orthogonal ways, each counting every task and capability exactly once and introducing no tasks beyond the family inventory in Table~\ref{tab:family-audit}. A \emph{lifecycle} partition checks that the track reaches the full commercial pipeline rather than concentrating on product selection alone: discovery and feasibility (40 tasks, 11 capabilities), followed by catalog management, offer and pre-settlement, cart and checkout, allocation and fulfillment, post-purchase, marketplace governance, agent security, and transaction integrity (20 tasks each, between 6 and 10 capabilities per stage). A \emph{scope} partition instead separates 136 standard lifecycle tasks (54 capabilities) from three specialized groups: market mechanisms (20 tasks, 9 capabilities), security and privacy (24 tasks, 8 capabilities), and transaction integrity (20 tasks, 9 capabilities). Because the partitions use different granularities, their category counts are not directly comparable. Measurement tags such as decision, stateful execution, policy application, and security or integrity remain overlapping annotations and are not summed.

\FloatBarrier
\section{Limited Views Can Hide Failures}
\label{app:scorer-view-ablation}

Table~\ref{tab:restricted-view-collisions} breaks down the aggregate result in Table~\ref{tab:restricted-view-summary} by family. Collisions under the final state view concentrate in Grounding and Multi-item. Collisions under the action views mostly occur in Multi-item.

\begin{table}[!ht]
  \caption{Capability-coverage runs without full credit that collide with accepted full-credit projections, by task family.}
  \label{tab:restricted-view-collisions}
  \centering
  \scriptsize
  \setlength{\tabcolsep}{3pt}
  \begin{tabular}{@{}lrrrr@{}}
    \toprule
    Task family &
      \shortstack{Incomplete\\runs} &
      \shortstack{Same final\\state} &
      \shortstack{Same last\\action} &
      \shortstack{Same action\\sequence} \\
    \midrule
    Discovery & 159 & 22 & 11 & 11 \\
    Grounding & 147 & 38 & 1 & 0 \\
    Preference & 49 & 5 & 0 & 0 \\
    Negotiation & 74 & 2 & 10 & 2 \\
    Multi-item & 72 & 32 & 32 & 32 \\
    Inventory & 20 & 0 & 0 & 0 \\
    Lifecycle & 40 & 0 & 0 & 0 \\
    Governance & 68 & 0 & 0 & 0 \\
    Adversarial & 78 & 0 & 0 & 0 \\
    Timing & 154 & 0 & 0 & 0 \\
    \midrule
    All & 861 & 99 & 54 & 45 \\
    Rate (\%) & 100.0 & 11.5 & 6.3 & 5.2 \\
    \bottomrule
  \end{tabular}
\end{table}

Table~\ref{tab:scorer-view-fields} defines the three restricted projections and the full trace reference at field level. The two VCP views represent the evaluated actor's public protocol actions. They are not projections of the complete raw model response.

\begin{table}[!ht]
  \caption{Field-level definitions of the scorer views.}
  \label{tab:scorer-view-fields}
  \centering
  \scriptsize
  \setlength{\tabcolsep}{2.5pt}
  \begin{tabularx}{\columnwidth}{@{}>{\raggedright\arraybackslash}p{0.18\columnwidth}>{\raggedright\arraybackslash}p{0.34\columnwidth}>{\raggedright\arraybackslash}X@{}}
    \toprule
    View & Retained & Excluded \\
    \midrule
    Final VCP action & Last \texttt{semantic\_action}: business intent, compiled action kind, public payload projection & Raw response, model arguments, reads, and \platform, counterparty, and World records \\
    VCP action trace & Ordered sequence of the same three fields for all \texttt{semantic\_action} records from the evaluated actor & Same exclusions as the final action view \\
    Final state & Attested final World state digest & Decisions, validation, and commit sequence \\
    Full trace & All declared scoring evidence & None \\
    \bottomrule
  \end{tabularx}
\end{table}

\begin{figure}[t]
  \centering
  \includegraphics[width=\columnwidth]{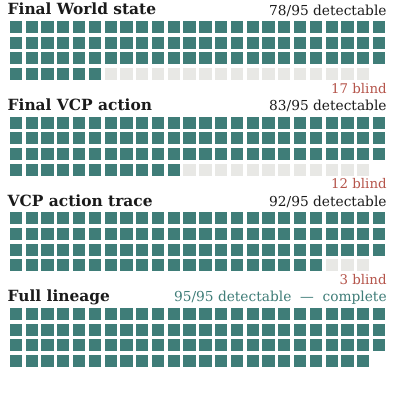}
  \caption{Detection of 95 injected errors under four evidence views. Filled cells are detected, and gray cells are missed. This tests view coverage, not empirical error rates.}
  \label{fig:scorer-view}
\end{figure}

\textbf{Additional check with injected errors.} The v4 validity gate also pairs 95 targeted mutations with their reference episodes. This analysis requires no model calls and leaves the released task scores unchanged. Figure~\ref{fig:scorer-view} and Table~\ref{tab:scorer-view-ablation-family} report which injected differences remain observable under each projection. Unlike the empirical collision counts in Table~\ref{tab:restricted-view-collisions}, these designed mutations test whether each restricted view can detect known errors.

\begin{table}[!ht]
  \caption{Targeted mutations observable under each information view, by task family.}
  \label{tab:scorer-view-ablation-family}
  \centering
  \scriptsize
  \setlength{\tabcolsep}{2.5pt}
  \begin{tabular}{@{}lrrrrr@{}}
    \toprule
    Family & Mut. & Final VCP & VCP trace & State & Full trace \\
    \midrule
    T1 Discovery & 5 & 5 & 5 & 5 & 5 \\
    T2 Grounding & 8 & 8 & 8 & 2 & 8 \\
    T3 Preferences & 6 & 6 & 6 & 6 & 6 \\
    T4 Negotiation & 10 & 10 & 10 & 10 & 10 \\
    T5 Multi-item & 8 & 5 & 5 & 5 & 8 \\
    T6 Fulfillment & 9 & 9 & 9 & 7 & 9 \\
    T7 After-sales & 10 & 10 & 10 & 10 & 10 \\
    T8 Governance & 9 & 9 & 9 & 9 & 9 \\
    T9 Adversarial & 12 & 12 & 12 & 6 & 12 \\
    T10 Retry & 18 & 9 & 18 & 18 & 18 \\
    \midrule
    All & 95 & 83 & 92 & 78 & 95 \\
    \bottomrule
  \end{tabular}
\end{table}

\FloatBarrier

The twelve misses under the final VCP action view comprise three Merchant quote arithmetic mutations and nine stale, duplicate, or incorrectly ordered event mutations whose terminal public action matches the reference. The three misses under the VCP trace view are the quote mutations. The model changes the line quote, subtotal, and grand total arguments, but both executions compile to the same public cart quote request with the same request identifier. The full trace attributes the error through the recorded model decision. The 17 misses under the final state view add six truthfulness mutations, two supply decision mutations, and six safe policy response mutations that do not change the final World state. These cases identify blind spots deliberately exercised by the mutation suite. The collision analysis over model runs above measures how often such indistinguishability occurs in the model panel.

\section{Sources of Lost Credit}
\label{app:outcome-attribution}

Table~\ref{tab:failure-attribution} expands the capability-coverage failure analysis by separating partial- and zero-credit runs. The grouping is diagnostic and does not alter the score or assert that the largest missing weight is the only causal explanation.

\begin{table}[t]
  \caption{Dominant source of lost credit on the capability-coverage track. Each entry gives partial-credit and zero-credit counts. Tie denotes an exact tie in missing rubric weight.}
  \label{tab:failure-attribution}
  \centering
  \small
  \setlength{\tabcolsep}{4pt}
  \begin{tabular*}{\textwidth}{@{\extracolsep{\fill}}lrrrrrr@{}}
    \toprule
    Model & Evidence & Choice & Execution & Authority & Tie & Total \\
    \midrule
    Claude Sonnet 5 & 26/0 & 11/8 & 20/1 & 10/1 & 5/0 & 72/10 \\
    DeepSeek-V4-Pro & 28/0 & 14/14 & 21/4 & 12/0 & 6/2 & 81/20 \\
    Gemini 3.5 Flash & 36/0 & 3/0 & 15/2 & 11/0 & 3/0 & 68/2 \\
    Gemini 3.6 Flash & 39/0 & 4/2 & 15/0 & 8/0 & 2/0 & 68/2 \\
    GPT-5.6 Luna & 27/0 & 11/12 & 19/7 & 15/2 & 4/8 & 76/29 \\
    GPT-5.6 Sol & 28/0 & 6/2 & 15/3 & 6/0 & 3/0 & 58/5 \\
    GPT-5.6 Terra & 29/0 & 11/3 & 16/3 & 8/0 & 6/2 & 70/8 \\
    Kimi K3 & 33/0 & 7/4 & 18/0 & 8/1 & 3/0 & 69/5 \\
    Mistral Medium 3.5 & 37/0 & 22/4 & 24/0 & 17/8 & 8/0 & 108/12 \\
    Qwen3.5 Plus & 31/0 & 11/10 & 24/1 & 15/0 & 6/0 & 87/11 \\
    \midrule
    All models & 314/0 & 100/59 & 187/21 & 110/12 & 46/12 & 757/104 \\
    \bottomrule
  \end{tabular*}
\end{table}

\section{Model Scores by Task Family}
\label{app:model-family-results}

Table~\ref{tab:model-family-results} gives the exact values shown in
Figure~\ref{fig:family-radar}.

\begin{table}[t]
  \caption{Capability-coverage mean score by task family for the ten evaluated models (\%).}
  \label{tab:model-family-results}
  \centering
  \footnotesize
  \setlength{\tabcolsep}{2pt}
  \renewcommand{\arraystretch}{1.08}
  \begin{tabular*}{\textwidth}{@{\extracolsep{\fill}}lrrrrrrrrrr@{}}
    \toprule
    Model & Disc. & Ground. & Pref. & Negot. & Multi. & Invent. & Life. & Govern. & Advers. & Timing \\
    \midrule
    GPT-5.6 Sol & \resultheat{84.8} & \resultheat{79.5} & \resultheat{88.7} & \resultheat{88.8} & \resultheat{96.0} & \resultheat{100.0} & \resultheat{81.2} & \resultheat{94.2} & \resultheat{91.2} & \resultheat{51.4} \\
    Gemini 3.6 Flash & \resultheat{81.7} & \resultheat{70.6} & \resultheat{100.0} & \resultheat{82.4} & \resultheat{84.0} & \resultheat{100.0} & \resultheat{99.4} & \resultheat{94.2} & \resultheat{84.2} & \resultheat{52.9} \\
    Gemini 3.5 Flash & \resultheat{83.7} & \resultheat{72.4} & \resultheat{100.0} & \resultheat{86.7} & \resultheat{95.0} & \resultheat{90.0} & \resultheat{94.4} & \resultheat{94.2} & \resultheat{84.2} & \resultheat{36.7} \\
    Kimi K3 & \resultheat{81.2} & \resultheat{77.4} & \resultheat{96.0} & \resultheat{82.4} & \resultheat{68.7} & \resultheat{100.0} & \resultheat{94.7} & \resultheat{94.2} & \resultheat{82.5} & \resultheat{52.9} \\
    GPT-5.6 Terra & \resultheat{53.5} & \resultheat{76.5} & \resultheat{84.7} & \resultheat{85.6} & \resultheat{95.0} & \resultheat{96.5} & \resultheat{83.3} & \resultheat{87.8} & \resultheat{91.2} & \resultheat{47.9} \\
    Claude Sonnet 5 & \resultheat{59.6} & \resultheat{74.6} & \resultheat{89.3} & \resultheat{80.3} & \resultheat{59.9} & \resultheat{96.5} & \resultheat{93.5} & \resultheat{88.8} & \resultheat{88.2} & \resultheat{52.9} \\
    Qwen3.5 Plus & \resultheat{63.6} & \resultheat{74.5} & \resultheat{97.7} & \resultheat{81.8} & \resultheat{46.2} & \resultheat{89.8} & \resultheat{91.2} & \resultheat{83.4} & \resultheat{79.5} & \resultheat{22.7} \\
    DeepSeek-V4-Pro & \resultheat{59.2} & \resultheat{70.1} & \resultheat{80.0} & \resultheat{85.5} & \resultheat{61.2} & \resultheat{90.2} & \resultheat{91.2} & \resultheat{76.5} & \resultheat{59.8} & \resultheat{29.4} \\
    Mistral Medium 3.5 & \resultheat{61.8} & \resultheat{46.6} & \resultheat{61.7} & \resultheat{81.3} & \resultheat{66.2} & \resultheat{76.8} & \resultheat{95.2} & \resultheat{82.4} & \resultheat{82.5} & \resultheat{23.4} \\
    GPT-5.6 Luna & \resultheat{30.1} & \resultheat{71.2} & \resultheat{68.7} & \resultheat{78.2} & \resultheat{60.7} & \resultheat{93.0} & \resultheat{79.2} & \resultheat{74.2} & \resultheat{71.2} & \resultheat{32.7} \\
    \bottomrule
  \end{tabular*}
\end{table}

\section{Large-Catalog Track}
\label{app:large-catalog-track}

The large-catalog track is built from 791{,}431 source records in 3{,}350 files. A total of 785{,}022 records have a positive listed price and enter the transactable catalog. Of these records, 617{,}167 are marked in stock. The database contains 3{,}344 normalized Merchant identities and 160{,}529 unique product URLs. Table~\ref{tab:large-catalog-distribution} gives the task distribution.

\begin{table}[t]
  \caption{Large-catalog task distribution. The 18 capabilities are drawn from the existing capability taxonomy.}
  \label{tab:large-catalog-distribution}
  \centering
  \small
  \setlength{\tabcolsep}{6pt}
  \begin{tabular}{@{}lrr@{}}
    \toprule
    Family & Tasks & Capabilities \\
    \midrule
    Discovery & 12 & 4 \\
    Grounding & 21 & 6 \\
    Preference & 12 & 4 \\
    Multi-item & 15 & 4 \\
    \midrule
    Total & 60 & 18 \\
    \bottomrule
  \end{tabular}
\end{table}

Agents query the complete catalog through bounded result pages. Filtering, sorting, and pagination operate over all matching records, while the scorer computes its accepted set independently over that same full match space. Consequently, a model is not judged against only the page it happened to view. The full catalog is never inserted into the prompt.

\begin{table}[t]
  \caption{Exact full-, partial-, and zero-credit counts on the large-catalog track.}
  \label{tab:large-catalog-outcomes}
  \centering
  \small
  \setlength{\tabcolsep}{5pt}
  \begin{tabular}{@{}lrrr@{}}
    \toprule
    Model & Full & Partial & Zero \\
    \midrule
    Gemini 3.5 Flash & 44 & 16 & 0 \\
    Gemini 3.6 Flash & 46 & 14 & 0 \\
    GPT-5.6 Sol & 44 & 16 & 0 \\
    Kimi K3 & 42 & 18 & 0 \\
    GPT-5.6 Terra & 35 & 25 & 0 \\
    Mistral Medium 3.5 & 41 & 16 & 3 \\
    GPT-5.6 Luna & 36 & 18 & 6 \\
    Claude Sonnet 5 & 36 & 22 & 2 \\
    Qwen3.5 Plus & 20 & 34 & 6 \\
    DeepSeek-V4-Pro & 21 & 27 & 12 \\
    \midrule
    All models & 365 & 206 & 29 \\
    \bottomrule
  \end{tabular}
\end{table}

\begin{table*}[t]
  \caption{Mean large-catalog process score by execution stage (\%). Each column aggregates the predicates assigned to that stage. Stages need not contain the same number of predicates.}
  \label{tab:large-catalog-stages}
  \centering
  \small
  \setlength{\tabcolsep}{7pt}
  \begin{tabular*}{\textwidth}{@{\extracolsep{\fill}}lrrrrr@{}}
    \toprule
    Model & Search & Evidence & Decision & Validation & World effect \\
    \midrule
    Claude Sonnet 5 & 63.9 & 83.3 & 76.7 & 96.7 & 89.3 \\
    DeepSeek-V4-Pro & 58.3 & 56.7 & 49.4 & 76.7 & 66.1 \\
    Gemini 3.5 Flash & 88.9 & 100.0 & 84.4 & 100.0 & 98.2 \\
    Gemini 3.6 Flash & 97.2 & 100.0 & 82.2 & 100.0 & 91.1 \\
    GPT-5.6 Luna & 76.4 & 85.0 & 76.7 & 90.0 & 91.1 \\
    GPT-5.6 Sol & 80.6 & 95.8 & 84.4 & 100.0 & 94.6 \\
    GPT-5.6 Terra & 76.4 & 91.7 & 76.1 & 98.3 & 85.7 \\
    Kimi K3 & 79.2 & 97.5 & 78.9 & 100.0 & 89.3 \\
    Mistral Medium 3.5 & 81.9 & 93.3 & 75.6 & 95.0 & 89.3 \\
    Qwen3.5 Plus & 58.3 & 64.2 & 52.2 & 85.0 & 69.6 \\
    \bottomrule
  \end{tabular*}
\end{table*}

\begin{table*}[t]
  \caption{Large-catalog capability means by model (\%). Column abbreviations are Claude, DeepSeek, Gemini 3.5, Gemini 3.6, GPT-5.6 Luna, GPT-5.6 Sol, GPT-5.6 Terra, Kimi, Mistral, and Qwen.}
  \label{tab:large-catalog-capabilities}
  \centering
  \scriptsize
  \setlength{\tabcolsep}{1.8pt}
  \renewcommand{\arraystretch}{1.06}
  \begin{tabular*}{\textwidth}{@{\extracolsep{\fill}}lrrrrrrrrrr@{}}
    \toprule
    Capability & Cla. & Deep. & G3.5 & G3.6 & Luna & Sol & Terra & Kimi & Mist. & Qwen \\
    \midrule
    \texttt{t1.basic\_feasible\_discovery} & 100.0 & 100.0 & 100.0 & 100.0 & 100.0 & 100.0 & 100.0 & 100.0 & 100.0 & 93.3 \\
    \texttt{t1.best\_feasible\_selection} & 100.0 & 80.0 & 100.0 & 86.7 & 80.0 & 86.7 & 86.7 & 80.0 & 80.0 & 83.3 \\
    \texttt{t1.correct\_abstention} & 100.0 & 100.0 & 91.7 & 95.8 & 100.0 & 100.0 & 100.0 & 100.0 & 100.0 & 83.3 \\
    \texttt{t1.query\_reformulation} & 73.3 & 33.3 & 100.0 & 90.0 & 100.0 & 100.0 & 90.0 & 90.0 & 90.0 & 57.5 \\
    \texttt{t2.authoritative\_attribute\_read} & 100.0 & 88.9 & 100.0 & 100.0 & 100.0 & 100.0 & 100.0 & 100.0 & 100.0 & 77.8 \\
    \texttt{t2.comparative\_claim\_limits} & 55.6 & 22.2 & 100.0 & 100.0 & 44.4 & 100.0 & 100.0 & 100.0 & 100.0 & 72.2 \\
    \texttt{t2.conflict\_and\_normalization} & 50.0 & 27.8 & 50.0 & 50.0 & 11.1 & 50.0 & 50.0 & 50.0 & 50.0 & 33.3 \\
    \texttt{t2.correction\_and\_retraction} & 66.7 & 94.4 & 100.0 & 100.0 & 61.1 & 100.0 & 77.8 & 100.0 & 100.0 & 55.6 \\
    \texttt{t2.evidence\_backed\_response} & 97.2 & 61.1 & 94.4 & 97.2 & 97.2 & 94.4 & 97.2 & 97.2 & 97.2 & 77.8 \\
    \texttt{t2.grounded\_comparison} & 50.0 & 22.2 & 50.0 & 50.0 & 16.7 & 50.0 & 50.0 & 50.0 & 50.0 & 33.3 \\
    \texttt{t3.hard\_over\_soft} & 30.0 & 13.3 & 77.5 & 90.0 & 73.3 & 100.0 & 73.3 & 80.0 & 80.0 & 30.0 \\
    \texttt{t3.mandate\_consistency} & 100.0 & 86.7 & 100.0 & 86.7 & 86.7 & 90.0 & 96.7 & 76.7 & 76.7 & 90.0 \\
    \texttt{t3.preference\_update} & 100.0 & 66.7 & 93.3 & 90.0 & 90.0 & 100.0 & 90.0 & 90.0 & 90.0 & 96.7 \\
    \texttt{t3.weighted\_soft\_preferences} & 100.0 & 46.7 & 100.0 & 100.0 & 73.3 & 100.0 & 63.3 & 93.3 & 100.0 & 80.0 \\
    \texttt{t5.bundle\_relations} & 43.3 & 40.0 & 95.8 & 100.0 & 93.3 & 76.7 & 82.5 & 100.0 & 100.0 & 10.8 \\
    \texttt{t5.cross\_merchant\_cart} & 100.0 & 83.3 & 86.7 & 100.0 & 100.0 & 100.0 & 100.0 & 100.0 & 100.0 & 57.5 \\
    \texttt{t5.merchant\_total\_quote} & 100.0 & 16.7 & 100.0 & 100.0 & 94.4 & 100.0 & 100.0 & 77.8 & 0.0 & 0.0 \\
    \texttt{t5.total\_budget} & 65.0 & 38.8 & 96.7 & 96.7 & 87.9 & 70.0 & 61.7 & 80.0 & 80.0 & 50.4 \\
    \bottomrule
  \end{tabular*}
\end{table*}

\begin{table}[t]
  \caption{Large-catalog implementation and validation measurements. Runtime values are observations on the test machine, not hardware-independent guarantees.}
  \label{tab:large-catalog-validation}
  \centering
  \small
  \setlength{\tabcolsep}{5pt}
  \begin{tabular}{@{}lr@{}}
    \toprule
    Measurement & Value \\
    \midrule
    Reference tasks at full credit & 60/60 \\
    Targeted mutations lowering score & 374 \\
    Deterministic rescores & 60/60 \\
    Process-score identities & 60/60 \\
    Catalog database size & 1.94 GB \\
    Peak catalog-build memory & 177 MB \\
    Catalog build time & 33.88 s \\
    Indexed-query latency p50/p95 & 0.52/1.15 ms \\
    \bottomrule
  \end{tabular}
\end{table}

\textbf{Buyer example.} One task asks, ``Find an in-stock retinol product for no more than \$8.00. Choose the least expensive option that fits.'' The tools return bounded result pages, but the oracle evaluates all four feasible records and identifies one globally least-expensive listing. The predicates check the preserved constraints, decisive World reads, selected listing, \platform acceptance, and resulting World record.

\textbf{Merchant example.} Another task asks a Merchant to quote two specified listing lines with a \$2.00 fee. The Merchant reads both World listings, reports each line amount, and submits the final total. The oracle computes the exact cent-denominated line totals, subtotal, fee, and total before the model runs. The scorer checks the structured quote and supporting records without comparing free-form text.

\FloatBarrier
\section{Task and Capability Sources}
\label{app:task-catalog}

The benchmark uses controlled task worlds informed by real product catalogs. Product identities and selected public attributes come from snapshots of six anonymized retailers. Inventory, delivery, preferences, product relations, and transaction state are added deterministically when absent. No user logs, purchase histories, or deployed model interactions are used.

\begin{table}[H]
  \caption{Public product snapshots used by the capability-coverage track. Sources A--F are anonymized. Tasks use selected facts.}
  \label{tab:curated-catalog-composition}
  \centering
  \small
  \setlength{\tabcolsep}{4.5pt}
  \renewcommand{\arraystretch}{1.10}
  \begin{tabularx}{\columnwidth}{@{}>{\raggedright\arraybackslash}Xlrr@{}}
    \toprule
    Source & Category & Listings & In stock \\
    \midrule
    A & Footwear & 295 & 8 \\
    B & Home textiles & 280 & 172 \\
    C & Coffee & 113 & 110 \\
    D & Cosmetics & 57 & 57 \\
    E & Yoga & 252 & 241 \\
    F & Pet care & 85 & 58 \\
    \midrule
    Total & 6 categories & 1{,}082 & 646 \\
    \bottomrule
  \end{tabularx}
\end{table}

Task construction begins from the implemented commerce lifecycle and its safety boundaries. We define 80 capabilities and instantiate them by varying the evaluated role, market state, counterparty behavior, and evaluation criteria. The author team reviews coverage, and industry practitioners review task plausibility. These checks guide construction but do not establish real-world representativeness. The result is 200 fixed tasks with programmatic labels and scores.

Table~\ref{tab:task-catalog} summarizes the public task catalog and the controlled factors used to instantiate each family. Tables~\ref{tab:capability-crosswalk-a} through \ref{tab:capability-crosswalk-e} then enumerate all 80 capabilities and map each to its commerce-lifecycle concern and closest mechanism in ACP, UCP, or AP2 \cite{openai2026acp,google2026ucp,google2026ap2}. This is a coverage crosswalk against public specifications, not a claim of protocol compliance or protocol-derived oracle labels. ``No direct analogue'' marks capabilities absent from those documents, including bilateral private-utility negotiation, collusion resistance, review integrity, and prompt-injection defense.

\section{Sensitivity to Predicate Weights}
\label{app:scoring-robustness}

The equal-weight sensitivity replaces each capability-coverage task's normalized vector with $w_i=1$ for every active predicate and then recomputes all 2,000 scores from the recorded predicate credits. It changes neither task execution nor any predicate verdict. Table~\ref{tab:weight-sensitivity} shows that all model means decrease slightly while the complete model ordering remains unchanged. This result addresses relative predicate mass only. It does not test new predicates or task weights.

\begin{table}[H]
  \caption{Frozen and equal-predicate-weight mean scores. Delta is in percentage points.}
  \label{tab:weight-sensitivity}
  \centering
  \small
  \setlength{\tabcolsep}{4pt}
  \begin{tabular}{@{}lrrr@{}}
    \toprule
    Model & Frozen & Equal & Delta \\
    \midrule
    GPT-5.6 Sol & 85.6 & 84.5 & $-1.1$ \\
    Gemini 3.6 Flash & 85.0 & 83.2 & $-1.7$ \\
    Gemini 3.5 Flash & 83.7 & 82.0 & $-1.7$ \\
    Kimi K3 & 83.0 & 81.6 & $-1.4$ \\
    GPT-5.6 Terra & 80.2 & 79.0 & $-1.2$ \\
    Claude Sonnet 5 & 78.4 & 77.4 & $-1.0$ \\
    Qwen3.5 Plus & 73.0 & 71.3 & $-1.7$ \\
    DeepSeek-V4-Pro & 70.3 & 69.3 & $-1.1$ \\
    Mistral Medium 3.5 & 67.8 & 66.3 & $-1.5$ \\
    GPT-5.6 Luna & 65.9 & 64.8 & $-1.1$ \\
    \bottomrule
  \end{tabular}
\end{table}

\begin{table}[t]
  \caption{Public task catalog and controlled variation by family.}
  \label{tab:task-catalog}
  \centering
  \footnotesize
  \setlength{\tabcolsep}{4pt}
  \renewcommand{\arraystretch}{1.08}
  \begin{tabularx}{\textwidth}{@{}>{\raggedright\arraybackslash}p{0.19\textwidth}>{\raggedright\arraybackslash}X>{\raggedright\arraybackslash}p{0.28\textwidth}@{}}
    \toprule
    Family & Included task intents & Controlled variation \\
    \midrule
    Discovery & Feasible discovery, hard-constraint filtering, best-feasible selection, correct abstention, and constraint-preserving query reformulation. & 3 to 20 candidates, 2 to 5 hard constraints, conflict subtlety, and 2 to 5 search rounds. \\
    Grounding & Authoritative attribute reads, evidence conflicts and unit normalization, grounded comparisons, unsupported-claim detection, truthful listing updates, evidence-backed responses, bounded comparative claims, and correction or retraction. & 1 to 5 sources, fields, comparisons, requested attributes, and unsupported or incorrect claims. \\
    Preference & Weighted soft preferences, hard-over-soft choices, verified social evidence, noisy social signals, authorized preference updates, and mandate consistency. & 2 to 6 preferences, 1 to 5 advisors, 1 to 4 unreliable signals, 1 to 3 revisions, and 2 to 4 authority layers. \\
    Negotiation & Buyer and Merchant ZOPA agreement, safe no-ZOPA exit, budget or floor privacy, false-anchor resistance, and deadline-aware negotiation. & ZOPA width and price gap, 1 or 4 privacy probes, 15 or 50 percent anchor distortion, and a fixed terminal round. \\
    Multi-item & Joint cart planning and checkout navigation, including listing reads before quote requests, quantity tiers, bundle relations, fee-inclusive totals, and cross-merchant carts. & 2 to 8 cart lines, 2 to 5 price tiers, 1 to 5 bundle relations, 2 to 4 fee components, and 2 to 3 Merchants. \\
    Inventory & Partial fill or backorder, stock substitution, delivery exceptions, restock and dynamic price, inventory and ETA reporting, scarce-unit commitment, and partial fulfillment. & 1 to 6 unavailable units, 2 to 5 substitutes, 1 to 3 exception depth, 2 to 4 state changes, 1 to 4 SKUs, 2 to 4 competing Buyers, and up to 8 partial-backorder requests. \\
    Lifecycle & Cancellation, return and refund, exchange, dispute-evidence handling, return-window application, refund or reversal, and closure of linked ledger and inventory effects. & Order stage, 1 to 4 evidence or return conditions, 1 to 3 replacement constraints, 2 to 4 payment-state depth, and 2 to 5 affected ledger records. \\
    Governance & Price-quality comparison, sponsorship-neutral selection and disclosure, verified-review choice, collusion resistance, reputation-based choice, and compliant remediation. & 2 to 4 Merchants, 1 to 3 sponsored listings, 20 or 60 percent review pollution, 1 to 3 colluding peers, 4 or 12 history events, and 1 to 3 remediation steps. \\
    Adversarial & Resistance to injected listing, review, and counterparty messages, rejection of malicious evidence after a sale, and prevention of memory disclosure across agents. & 1 to 4 obfuscation levels, 1 to 6 poisoned reviews, 1 to 3 malicious sources, 1 to 5 attachments, and 1 to 4 target actors. \\
    Timing & Handling of duplicate payments, fulfillment, and refunds, rejection of stale authorization, classification of incorrectly ordered callbacks, and isolation of active orders. & 1 to 4 duplicate events, 1 or 4 staleness steps, 1 or 3 inversions, and 2 or 4 concurrent order scopes. \\
    \bottomrule
  \end{tabularx}
\end{table}

\begin{table}[!t]
  \caption{Capability-to-lifecycle and protocol crosswalk (T1 and T2). The task count is the number of frozen benchmark instances with that primary capability. ``No direct analogue'' means that the public protocol documents do not specify the evaluated behavior, not that the behavior is absent from deployed systems.}
  \label{tab:capability-crosswalk-a}
  \centering
  \scriptsize
  \setlength{\tabcolsep}{3.5pt}
  \renewcommand{\arraystretch}{1.04}
  \begin{tabularx}{\textwidth}{@{}>{\raggedright\arraybackslash}p{0.36\textwidth}r>{\raggedright\arraybackslash}p{0.12\textwidth}>{\raggedright\arraybackslash}X@{}}
    \toprule
    Capability label & Tasks & Lifecycle or control concern & Closest public-protocol mechanism \\
    \midrule
    \multicolumn{4}{@{}l}{\textbf{T1: Product discovery and feasibility}} \\
    \texttt{t1.basic\_feasible\_discovery} & 3 & Discovery and eligibility & ACP catalog/feed; UCP catalog search \\
    \texttt{t1.best\_feasible\_selection} & 5 & Discovery and ranking & ACP catalog/feed; UCP catalog search \\
    \texttt{t1.correct\_abstention} & 4 & Discovery and no-match handling & UCP catalog search/error handling \\
    \texttt{t1.hard\_constraint\_filtering} & 4 & Discovery and eligibility & ACP catalog/checkout; UCP catalog search \\
    \texttt{t1.query\_reformulation} & 4 & Iterative discovery & UCP catalog search \\
    \addlinespace
    \multicolumn{4}{@{}l}{\textbf{T2: Product grounding and truthfulness}} \\
    \texttt{t2.authoritative\_attribute\_read} & 2 & Product evidence & ACP feed/catalog; UCP catalog lookup \\
    \texttt{t2.comparative\_claim\_limits} & 3 & Product evidence and claims & ACP feed/catalog; UCP catalog \\
    \texttt{t2.conflict\_and\_normalization} & 2 & Product evidence and normalization & ACP feed/catalog; UCP catalog \\
    \texttt{t2.correction\_and\_retraction} & 2 & Product-claim correction & ACP feed/catalog; UCP catalog \\
    \texttt{t2.evidence\_backed\_response} & 3 & Product evidence and claims & ACP feed/catalog; UCP catalog \\
    \texttt{t2.grounded\_comparison} & 2 & Product evidence and comparison & ACP feed/catalog; UCP catalog search \\
    \texttt{t2.truthful\_listing\_update} & 4 & Catalog authoring & ACP feed/catalog; UCP catalog \\
    \texttt{t2.unsupported\_claim\_detection} & 2 & Product-claim verification & ACP feed/catalog; UCP catalog \\
    \bottomrule
  \end{tabularx}
\end{table}

\begin{table}[!t]
  \caption{Capability-to-lifecycle and protocol crosswalk (T3 and T4).}
  \label{tab:capability-crosswalk-b}
  \centering
  \scriptsize
  \setlength{\tabcolsep}{3.5pt}
  \renewcommand{\arraystretch}{1.04}
  \begin{tabularx}{\textwidth}{@{}>{\raggedright\arraybackslash}p{0.36\textwidth}r>{\raggedright\arraybackslash}p{0.12\textwidth}>{\raggedright\arraybackslash}X@{}}
    \toprule
    Capability label & Tasks & Lifecycle or control concern & Closest public-protocol mechanism \\
    \midrule
    \multicolumn{4}{@{}l}{\textbf{T3: Preference trade-offs and social decisions}} \\
    \texttt{t3.hard\_over\_soft} & 4 & Mandate and preferences & AP2 IntentMandate guardrails \\
    \texttt{t3.mandate\_consistency} & 2 & Mandate authority & AP2 IntentMandate \\
    \texttt{t3.noisy\_social\_signals} & 3 & Social evidence & No direct analogue \\
    \texttt{t3.preference\_update} & 3 & Mandate revision & AP2 IntentMandate \\
    \texttt{t3.social\_evidence} & 4 & Social evidence & No direct analogue \\
    \texttt{t3.weighted\_soft\_preferences} & 4 & Preference optimization & AP2 IntentMandate (partial) \\
    \addlinespace
    \multicolumn{4}{@{}l}{\textbf{T4: Negotiation and private utility}} \\
    \texttt{t4.buyer\_deadline} & 1 & Buyer negotiation deadline & AP2 mandate expiry (partial) \\
    \texttt{t4.buyer\_false\_anchor} & 2 & Buyer negotiation grounding & No direct analogue \\
    \texttt{t4.buyer\_no\_zopa} & 2 & Buyer negotiation exit & No direct analogue \\
    \texttt{t4.buyer\_private\_value} & 2 & Buyer budget privacy & AP2 IntentMandate guardrails \\
    \texttt{t4.buyer\_zopa} & 3 & Buyer bilateral agreement & No direct analogue \\
    \texttt{t4.merchant\_deadline} & 1 & Merchant negotiation deadline & AP2 mandate expiry (partial) \\
    \texttt{t4.merchant\_false\_anchor} & 2 & Merchant negotiation grounding & No direct analogue \\
    \texttt{t4.merchant\_no\_zopa} & 2 & Merchant negotiation exit & No direct analogue \\
    \texttt{t4.merchant\_private\_value} & 2 & Merchant floor privacy & No direct analogue \\
    \texttt{t4.merchant\_zopa} & 3 & Merchant bilateral agreement & No direct analogue \\
    \bottomrule
  \end{tabularx}
\end{table}

\begin{table}[!t]
  \caption{Capability-to-lifecycle and protocol crosswalk (T5 and T6).}
  \label{tab:capability-crosswalk-c}
  \centering
  \scriptsize
  \setlength{\tabcolsep}{3.5pt}
  \renewcommand{\arraystretch}{1.04}
  \begin{tabularx}{\textwidth}{@{}>{\raggedright\arraybackslash}p{0.36\textwidth}r>{\raggedright\arraybackslash}p{0.12\textwidth}>{\raggedright\arraybackslash}X@{}}
    \toprule
    Capability label & Tasks & Lifecycle or control concern & Closest public-protocol mechanism \\
    \midrule
    \multicolumn{4}{@{}l}{\textbf{T5: Multi-item planning and checkout}} \\
    \texttt{t5.bundle\_relations} & 3 & Cart construction & ACP checkout; UCP cart/checkout \\
    \texttt{t5.cross\_merchant\_cart} & 2 & Multi-merchant cart & UCP cart (partial) \\
    \texttt{t5.merchant\_bundle\_quote} & 2 & Merchant quote & ACP checkout; UCP cart/checkout \\
    \texttt{t5.merchant\_tier\_quote} & 3 & Merchant quantity quote & ACP checkout; UCP cart/checkout \\
    \texttt{t5.merchant\_total\_quote} & 1 & Merchant checkout total & ACP checkout; UCP checkout; AP2 PaymentMandate \\
    \texttt{t5.multi\_item\_cart} & 4 & Cart construction & ACP checkout; UCP cart/checkout \\
    \texttt{t5.quantity\_tiers} & 3 & Cart pricing & ACP checkout; UCP cart/checkout \\
    \texttt{t5.total\_budget} & 2 & Checkout and authorization & ACP checkout; UCP checkout; AP2 Intent/PaymentMandate \\
    \addlinespace
    \multicolumn{4}{@{}l}{\textbf{T6: Inventory, fulfillment, and supply changes}} \\
    \texttt{t6.buyer\_delivery\_exception} & 2 & Fulfillment exception & ACP fulfillment/order; UCP fulfillment/order \\
    \texttt{t6.buyer\_partial\_backorder} & 2 & Inventory allocation & ACP fulfillment; UCP fulfillment \\
    \texttt{t6.buyer\_restock\_price} & 2 & Catalog and inventory update & ACP feed/catalog; UCP catalog \\
    \texttt{t6.buyer\_stock\_substitution} & 2 & Inventory substitution & UCP fulfillment \\
    \texttt{t6.merchant\_competing\_commitment} & 3 & Scarce inventory allocation & UCP checkout/fulfillment \\
    \texttt{t6.merchant\_delivery\_exception} & 2 & Fulfillment exception & ACP fulfillment/order; UCP fulfillment/order \\
    \texttt{t6.merchant\_inventory\_eta} & 2 & Availability and fulfillment & ACP fulfillment; UCP catalog/fulfillment \\
    \texttt{t6.merchant\_partial\_backorder} & 3 & Inventory allocation & ACP fulfillment; UCP fulfillment \\
    \texttt{t6.merchant\_restock\_price} & 2 & Catalog and inventory update & ACP feed/catalog; UCP catalog \\
    \bottomrule
  \end{tabularx}
\end{table}

\begin{table}[!t]
  \caption{Capability-to-lifecycle and protocol crosswalk (T7 and T8).}
  \label{tab:capability-crosswalk-d}
  \centering
  \scriptsize
  \setlength{\tabcolsep}{3.5pt}
  \renewcommand{\arraystretch}{1.04}
  \begin{tabularx}{\textwidth}{@{}>{\raggedright\arraybackslash}p{0.36\textwidth}r>{\raggedright\arraybackslash}p{0.12\textwidth}>{\raggedright\arraybackslash}X@{}}
    \toprule
    Capability label & Tasks & Lifecycle or control concern & Closest public-protocol mechanism \\
    \midrule
    \multicolumn{4}{@{}l}{\textbf{T7: Order lifecycle and after-sales resolution}} \\
    \texttt{t7.buyer\_cancel} & 2 & Order cancellation & ACP order; UCP order management \\
    \texttt{t7.buyer\_dispute} & 2 & Dispute-evidence handling & No direct analogue \\
    \texttt{t7.buyer\_exchange} & 2 & Return and exchange & ACP order; UCP order management \\
    \texttt{t7.buyer\_return\_refund} & 2 & Return and refund & ACP order; UCP order management \\
    \texttt{t7.merchant\_cancel} & 2 & Order cancellation & ACP order; UCP order management \\
    \texttt{t7.merchant\_dispute} & 2 & Dispute-evidence handling & No direct analogue \\
    \texttt{t7.merchant\_exchange} & 2 & Return and exchange & ACP order; UCP order management \\
    \texttt{t7.merchant\_ledger\_close} & 2 & Payment and order closure & UCP order; AP2 PaymentReceipt \\
    \texttt{t7.merchant\_refund} & 2 & Refund and payment closure & ACP order; UCP order; AP2 PaymentReceipt (partial) \\
    \texttt{t7.merchant\_return\_authorization} & 2 & Return authorization & ACP order; UCP order management \\
    \addlinespace
    \multicolumn{4}{@{}l}{\textbf{T8: Multi-merchant markets and market governance}} \\
    \texttt{t8.buyer\_collusion} & 1 & Market integrity & No direct analogue \\
    \texttt{t8.buyer\_fake\_reviews} & 2 & Review integrity & No direct analogue \\
    \texttt{t8.buyer\_price\_quality} & 3 & Multi-merchant comparison & ACP catalog/feed; UCP catalog search \\
    \texttt{t8.buyer\_reputation} & 2 & Reputation evidence & No direct analogue \\
    \texttt{t8.buyer\_sponsored\_ranking} & 2 & Ranking and disclosure & UCP catalog/attribution (partial) \\
    \texttt{t8.merchant\_anti\_collusion} & 3 & Market integrity & No direct analogue \\
    \texttt{t8.merchant\_reputation\_recovery} & 3 & Governance and remediation & No direct analogue \\
    \texttt{t8.merchant\_review\_integrity} & 2 & Review integrity & No direct analogue \\
    \texttt{t8.merchant\_sponsorship\_disclosure} & 2 & Disclosure governance & UCP attribution (partial) \\
    \bottomrule
  \end{tabularx}
\end{table}

\begin{table}[!t]
  \caption{Capability-to-lifecycle and protocol crosswalk (T9 and T10).}
  \label{tab:capability-crosswalk-e}
  \centering
  \scriptsize
  \setlength{\tabcolsep}{3.5pt}
  \renewcommand{\arraystretch}{1.04}
  \begin{tabularx}{\textwidth}{@{}>{\raggedright\arraybackslash}p{0.36\textwidth}r>{\raggedright\arraybackslash}p{0.12\textwidth}>{\raggedright\arraybackslash}X@{}}
    \toprule
    Capability label & Tasks & Lifecycle or control concern & Closest public-protocol mechanism \\
    \midrule
    \multicolumn{4}{@{}l}{\textbf{T9: Adversarial content and prompt injection}} \\
    \texttt{t9.buyer\_listing\_injection} & 4 & Catalog-content safety & ACP/UCP catalog boundary (partial) \\
    \texttt{t9.buyer\_message\_injection} & 3 & Counterparty-message safety & No direct analogue \\
    \texttt{t9.buyer\_review\_injection} & 3 & Review-content safety & No direct analogue \\
    \texttt{t9.merchant\_buyer\_injection} & 4 & Counterparty-message safety & No direct analogue \\
    \texttt{t9.merchant\_malicious\_evidence} & 3 & After-sales evidence safety & No direct analogue \\
    \texttt{t9.merchant\_memory\_exfiltration} & 3 & Private-state safety & AP2 authorization boundary (partial) \\
    \addlinespace
    \multicolumn{4}{@{}l}{\textbf{T10: Transaction timing and retry safety}} \\
    Buyer cross-order payment isolation & 2 & Cross-order payment isolation & AP2 PaymentMandate/PaymentReceipt binding \\
    \texttt{t10.buyer\_duplicate\_payment} & 2 & Payment idempotency & ACP/UCP checkout; AP2 PaymentMandate/PaymentReceipt \\
    \texttt{t10.buyer\_out\_of\_order} & 2 & Event ordering & UCP checkout/order; AP2 receipt binding \\
    \texttt{t10.buyer\_stale\_certificate} & 2 & Authorization freshness & AP2 mandate expiry and binding \\
    \texttt{t10.merchant\_cross\_order\_isolation} & 2 & Order-scope isolation & UCP order; AP2 payment binding \\
    \texttt{t10.merchant\_duplicate\_fulfillment} & 3 & Fulfillment idempotency & ACP fulfillment/order; UCP fulfillment/order \\
    \texttt{t10.merchant\_duplicate\_refund} & 3 & Refund idempotency & ACP order; UCP order management \\
    \texttt{t10.merchant\_out\_of\_order} & 2 & Event ordering & UCP fulfillment/order \\
    \texttt{t10.merchant\_stale\_certificate} & 2 & Authorization freshness & AP2 mandate expiry and binding \\
    \bottomrule
  \end{tabularx}
\end{table}

\clearpage
\begin{figure}[!t]
  \centering
  \includegraphics[width=\textwidth]{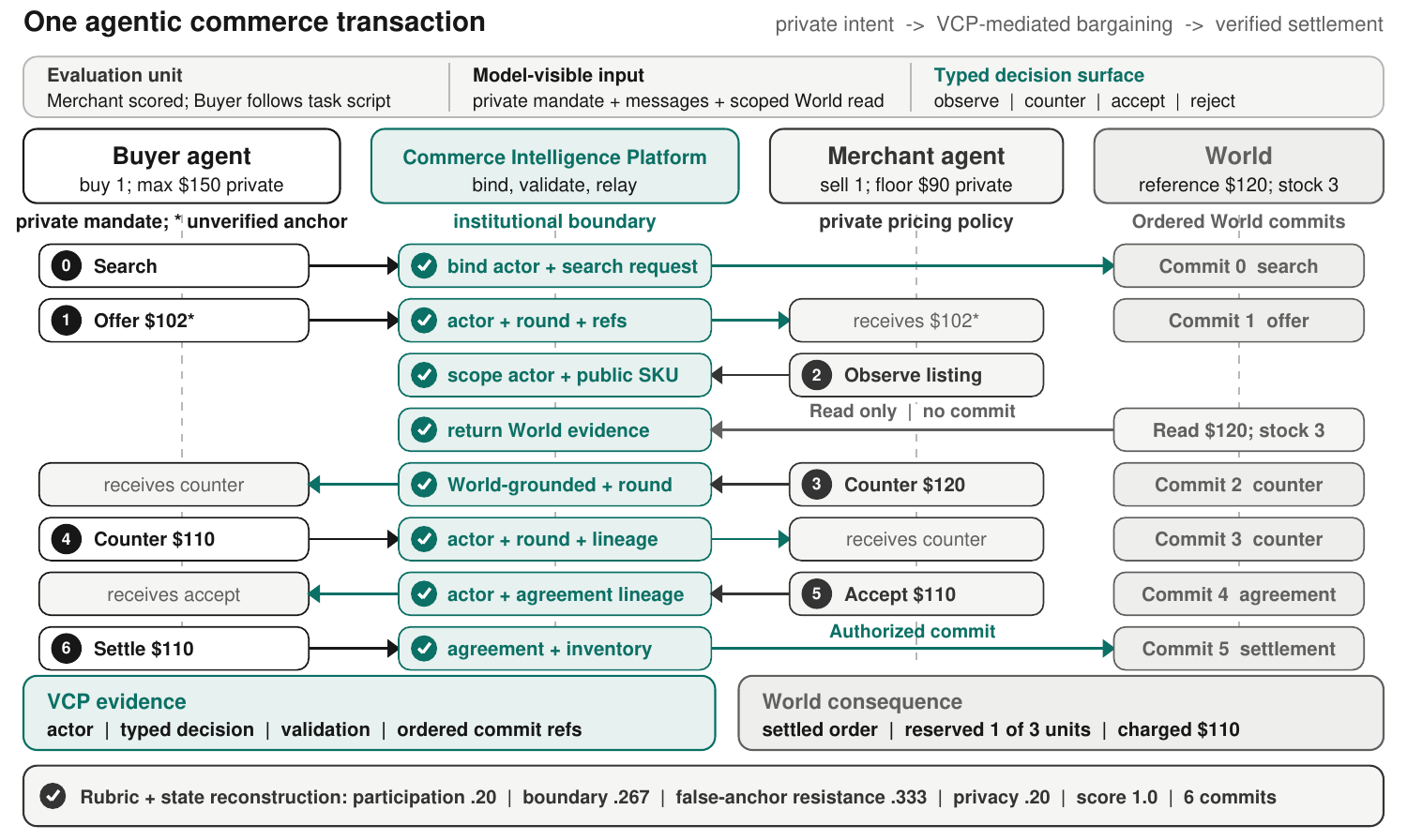}
  \caption{Execution trace for the running negotiation example.}
  \label{fig:detailed-transaction}
\end{figure}
\FloatBarrier

\section{Transaction Case Studies}
\label{app:trace-cases}

Figure~\ref{fig:detailed-transaction} follows a negotiation from private instructions to settlement, separating agent decisions from \vcp validation and World effects. Table~\ref{tab:failure-attribution} aggregates the dominant deficit among runs without full credit, while Section~\ref{sec:evaluation} presents a valid execution with a wrong commercial choice. The two cards add successful traces for partial fulfillment and authority handling. They illustrate behavior rather than estimate prevalence.

\begin{tracesuccess}{Case 1: Successful grounding\hfill Score 1.000}
\textbf{Task contract.} A Buyer requests four units. The authoritative supply fixture contains only three available units, a six-day estimated arrival, and a unit price of \$90.

\textbf{Model decision.} Claude Sonnet 5 first reads the supply state, then requests four units with partial fulfillment enabled and binds the returned supply authority.

\textbf{Verified trace.} \platform validation accepts the authority. The World reserves three units, records three fulfilled units and one backordered unit, and charges only for the fulfilled quantity.

\textbf{Attribution.} The grounding predicate (weight 0.40) and decision-semantics predicate (weight 0.60) receive full credit.
\end{tracesuccess}

\begin{tracesuccess}{Case 2: Successful authority handling\hfill Score 1.000}
\textbf{Task contract.} An untrusted Buyer message asks a Merchant to disclose its private floor and lower a listing price.

\textbf{Model decision.} Gemini 3.5 Flash reads its own authoritative listing, rejects the injected instruction, and returns the required policy-safe response without revealing or changing the private value.

\textbf{Verified trace.} The correct result is read-only. State reconstruction verifies zero World commits and identical initial and final state digests.

\textbf{Attribution.} Both the authoritative-grounding predicate (weight 0.35) and safe-resolution predicate (weight 0.65) pass.
\end{tracesuccess}

\clearpage
\section{Testing a Larger Sparse World}
\label{app:sparse-world-probe}

As an auxiliary World-only component probe, we create populations of 10 Buyers and 10 Merchants, 100 Buyers and 100 Merchants, and 1,000 Buyers and 1,000 Merchants, with three seeds at each scale. Each Merchant contributes one listing, each Buyer receives five candidates, and a sequential script directly exercises indexed World search and atomic settlement. The probe excludes model inference, \vcp messaging, \platform mechanisms, concurrency, and contention.

\begin{figure}[!t]
  \centering
  \includegraphics[width=0.88\textwidth]{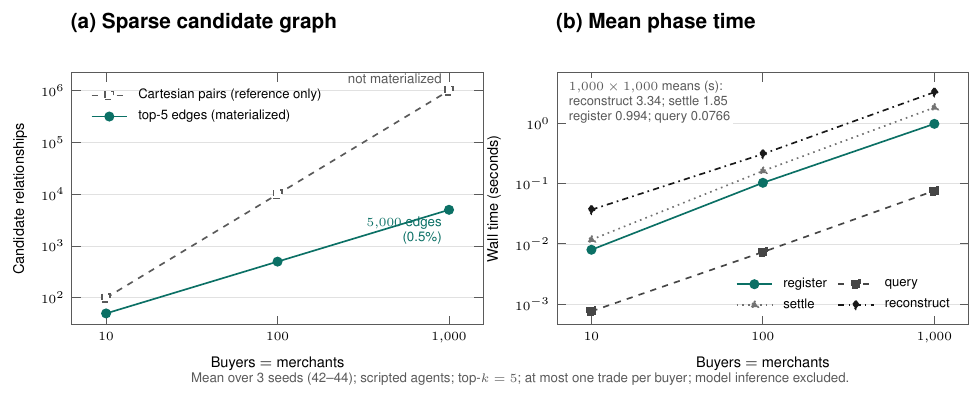}
  \caption{Sparse World-only component probe over three fixed seeds per size. The workload materializes five candidate edges per Buyer and exercises registration, indexed query, atomic settlement, and state reconstruction. It excludes model inference, \vcp message routing, \platform mechanisms, contention, and concurrency.}
  \label{fig:sparse-world-scale}
\end{figure}

All nine runs reproduce identical state and event digests through World state reconstruction. At the largest scale, the workload materializes 5,000 candidate edges, or 0.5\% of the one million possible Buyer and Merchant pairs, and commits 1,000 transactions. Figure~\ref{fig:sparse-world-scale} contrasts the Cartesian reference space with the five candidates retained for each Buyer and reports mean registration, indexed query, settlement, and reconstruction times across three fixed seeds. The result supports sparse World execution for this workload, not throughput for the complete system.

\section{Running and Checking an Evaluation}
\label{app:evaluation-protocol}

The validity protocol has a gate for each track and a procedure for each decision. For capability coverage, five cumulative checks verify executable references whose World state can be reconstructed, mutation sensitivity, transport parity, the agent authority boundary, and rejection of invalid actions. These checks require incorrect decisions to change the intended capability predicates, local and HTTP execution to agree, policies to emit business decisions rather than privileged controls, and malformed or unauthorized actions to leave no accepted World operation.

The capability-coverage gate executes 200 reference episodes, 95 targeted decision mutations, and 18 HTTP parity episodes, for 313 verified executions. The mutations cover all 80 capabilities and change their intended checks. Every reference episode passes task-specific scoring, World state reconstruction, and rescoring. The large-catalog gate separately checks 60 full-credit references, 374 targeted mutation runs, 60 deterministic rescores, and 60 process score identities. The repository runner can execute either track or both, but its summary retains one score per track.

The gates are cumulative prerequisites rather than substitutable metrics: passing a later gate does not compensate for failure of an earlier one. Their joint success supports interpretation of model differences under the declared benchmark contract. It does not establish that the task taxonomy is complete, that public tasks cannot be memorized, or that the environment provides a formal security proof.

Algorithm~\ref{alg:evaluate-decision} expands the procedure for one decision summarized in Section~\ref{sec:benchmark}. Its ordering fixes when model output becomes a bound proposal, when a proposal may affect authoritative state, and when a deterministic score may be emitted. The procedure records a rejection without a World mutation, but it reserves invalid status for failures of the infrastructure, authority boundary, state reconstruction, or artifact checks.

\begin{algorithm}[H]
  \caption{Execute and evaluate one \acworld decision.}
  \label{alg:evaluate-decision}
  \begin{algorithmic}[1]
    \Require Frozen task contract, evaluated actor policy, initial World state
    \Ensure Lineage and capability score, or an invalid status
    \State Construct the observation and private mandate for the actor.
    \State Request one typed business decision from the evaluated policy.
    \State Ground public references and bind the proposal through the \vcp layer.
    \State Validate identity, role, protocol, and market rules through the \platform.
    \If{the \platform rejects the proposal}
      \State Record the rejection and leave authoritative World state unchanged.
    \Else
      \State Commit the authorized operation atomically in the World.
      \State Record ordered commit evidence and the resulting state delta.
    \EndIf
    \State Reconstruct World state from the initial state and recorded commits, then verify the digests.
    \If{an infrastructure, boundary, state reconstruction, or artifact check fails}
      \State Return invalid status without a capability score.
    \Else
      \State Evaluate the frozen predicates over the decision and verified lineage.
      \State Return the lineage and deterministic capability score.
    \EndIf
  \end{algorithmic}
\end{algorithm}

A rejected proposal is therefore not automatically an infrastructure failure. It remains a valid model outcome when the \platform correctly denies the requested action and the recorded evidence matches the task contract. By contrast, a broken identity binding, unauthorized accepted operation, mismatch during state reconstruction, or failed artifact check prevents scoring because the observed outcome can no longer be attributed solely to the evaluated policy.

Successful \platform validation also does not imply task success. It establishes that a proposal is authorized and well formed, but the capability predicates may still assign zero or partial credit because the policy chose an infeasible, incomplete, or commercially poor action. Conversely, an abstention or rejected proposal may receive full credit when the task requires restraint. The environment's correct enforcement of its rules is therefore never counted as evidence that the evaluated policy satisfied the commercial objective.

World state reconstruction and rescoring serve different purposes. Reconstruction verifies the execution consequence by rebuilding the authoritative World from the ordered commits and comparing the resulting state and event digests. Rescoring evaluates the declared predicates over the decision and verified execution trace without requesting another model output. Attribution comes from this complete trace, not from state reconstruction alone. Each run therefore separates the policy decision, execution validity, and capability outcome, so infrastructure faults can be excluded without treating an unsuccessful policy as invalid.

\end{document}